\pdfoutput=1
\RequirePackage{float}
\documentclass{bmvc2k}

\title{Watch Your Speech: \\Text-aware Video-to-Speech Synthesis \\ with Textual Conditioning}
\addauthor{Gunwoo Lee}{gunwoo@soongsil.ac.kr}{1*}
\addauthor{Yoori Oh}{yoori0203@snu.ac.kr}{2*}
\addauthor{Yoseob Han}{yoseob.han@ssu.ac.kr}{3}
\addauthor{\scriptsize $^{*}$Equal contribution.}{}{}
\addinstitution{
 Department of Information and Telecommunication Engineering\\
 Soongsil University\\
 Seoul, Republic of Korea
}
\addinstitution{
 Graduate School of Data Science\\
 Seoul National University\\
 Seoul, Republic of Korea
}
\addinstitution{
 Department of Electronic Engineering\\
 Soongsil University\\
 Seoul, Republic of Korea
}

\usepackage{hyperref}
\usepackage{cleveref}
\usepackage{graphicx}
\usepackage{booktabs}
\usepackage[table]{xcolor}
\usepackage{multirow}
\usepackage{subcaption}
\usepackage{amsmath}
\usepackage{amssymb}
\usepackage{algorithm}
\usepackage{algorithmic}

\runninghead{Lee et al.}{Watch your speech}

\begin{document}

\maketitle

\begin{abstract}
Video-to-speech synthesis aims to generate natural-sounding speech from silent talking-face videos while ensuring phonetic accuracy. A fundamental challenge in this task is the inherent one-to-many mapping problem, where visual dynamics often lack sufficient information to uniquely determine the corresponding utterance. To address this, we propose \textit{Watch Your Speech} (WYS), a video-to-speech synthesis framework that incorporates textual conditioning as an explicit linguistic cue to mitigate visual ambiguity.
Our framework features an attention-based embedding fusion module that synergistically integrates textual context with video sequences, coupled with a conditional flow matching objective for high-fidelity speech generation. Extensive experiments on the LRS2 and LRS3 datasets demonstrate that WYS achieves superior performance, establishing new state-of-the-art results in audio-visual synchronization (LSE-C/D) while maintaining highly competitive textual accuracy (WER). Subjective evaluations further confirm that our model generates speech with near-human naturalness, validating the effectiveness of textual conditioning in content-controlled video-to-speech synthesis.

\noindent Project page: \url{https://github.com/gunwoo5034/Watch-your-Speech}
\end{abstract}

\section{Introduction}
\label{sec:intro}
In various real-world scenarios, such as low-quality video recordings, real-time video calls, or interactions with individuals who have speech impediments, audio can often become unintelligible or misaligned with the corresponding video.
Such degradation can result in a substantial loss of intelligibility and semantic content.
To mitigate these challenges, the field of video-to-speech (V2S) synthesis has emerged. The primary goal of V2S is to generate clear and natural-sounding speech from visual input alone, thereby restoring the intended sentence context.

\begin{figure}[t]
    \centering
    \includegraphics[width=0.8\textwidth]{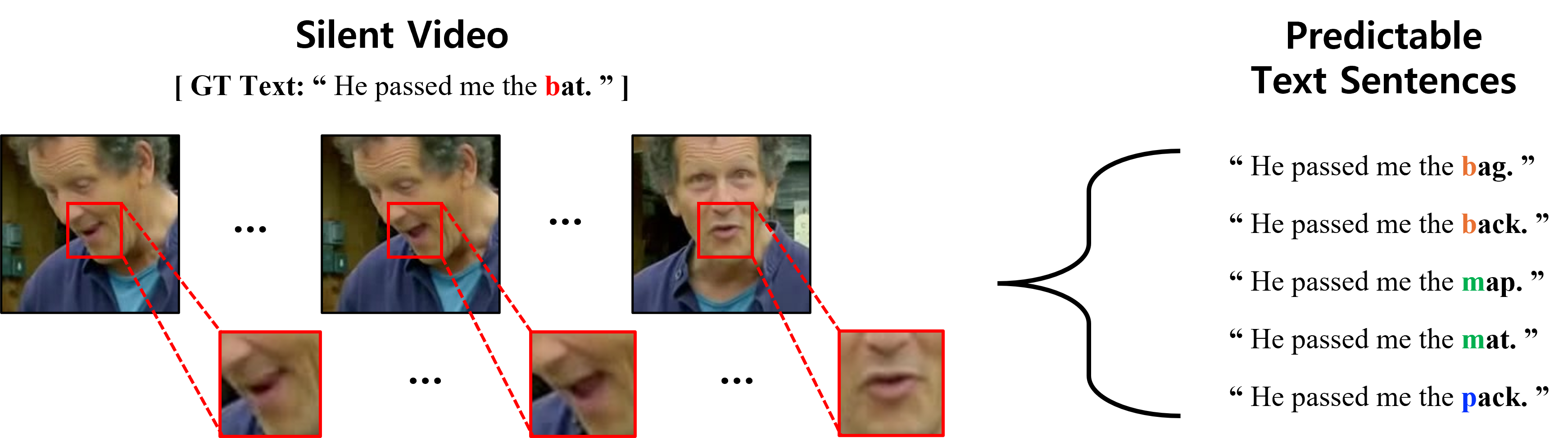}
    \vspace{5mm}
    \caption{The one-to-many mapping problem: visually similar lip movements for bilabial phonemes (/b/, /m/, /p/) lead to ambiguous speech predictions.}
    \label{fig:lip_movement}
\end{figure}

Previous approaches to synthesize realistic speech from silent talking-face video have leveraged the speaker's lip movements and facial expressions. However, these methods are fundamentally limited by the one-to-many mapping problem, where visually similar lip shapes can correspond to multiple distinct phonemes \cite{fernandez2018survey,fisher1968confusions}. This visual ambiguity makes it difficult for the synthesized speech to achieve high content accuracy. For instance, the lip movements for bilabial phonemes such as /b/ and /m/—as in ``bat'' and ``mat''—are often visually indistinguishable, as shown in Fig.~\ref{fig:lip_movement}.

To address this limitation, recent methods have sought to integrate richer contextual information into the generation process. SVTS~\cite{mira2022svts} and DiffV2S~\cite{Choi_2023_ICCV} leverage audio-visual pre-trained representations to synthesize speech, yet the generated audio still lacks accurate content due to insufficient utilization of textual information, which can leave the one-to-many ambiguity insufficiently addressed.
A more recent approach, LipVoicer~\cite{yemini2023lipvoicer}, injects textual context via classifier guidance from a lip-reading model during inference. However, since text information is absent during training, the model has limited opportunity to learn audio-textual alignments within the generation process, and its reliance on an external classifier can affect speech naturalness.

To mitigate this problem, we propose \textit{Watch Your Speech} (WYS), a V2S synthesis framework that incorporates explicit textual conditioning from the training phase.
By integrating Ground-Truth (GT) text during training and leveraging a pre-trained lip-reading model~\cite{ma2023auto} to estimate text during inference, WYS substantially narrows the mapping space, alleviating the one-to-many ambiguity.
To synergistically fuse this textual conditioning with visual features, WYS employs an attention-based embedding fusion module that computes a modality-aligned representation, which then serves as the conditional input for a Conditional Flow Matching (CFM) objective, enabling the generation of high-fidelity speech that faithfully reflects the spoken content.

Experiments on the LRS2~\cite{afouras2018a} and LRS3~\cite{afouras2018b} benchmarks demonstrate that WYS improves audio-visual synchronization~\cite{chung2016} and perceptual quality while maintaining competitive textual accuracy. Subjective evaluations further indicate improved perceived naturalness compared with baseline methods.

\medskip
The main contributions of the proposed WYS framework are as follows:
\begin{itemize}
\item A V2S framework that directly integrates textual conditioning from the training phase onward, utilizing GT text during training and text predicted by a lip-reading model during inference, without relying on external classifier guidance.
\item An effective attention-based fusion module that explicitly aligns video and text features, coupled with a CFM objective for high-fidelity speech generation, thereby significantly alleviating the inherent one-to-many ambiguity of the V2S task.
\item Demonstration of improved audio-visual synchronization (LSE-C/D) and perceptual quality on the LRS2 and LRS3 benchmarks while maintaining competitive textual accuracy (WER).
\end{itemize}

\section{Related Work}
\label{sec:Related}

The objective of V2S synthesis is to generate intelligible, natural-sounding speech that remains tightly synchronized with silent talking-face video. Early research in this field primarily leveraged convolutional neural networks to extract and map visual features to acoustic representations, such as in Vid2Speech~\cite{ephrat2017vid2speech} and LTBS~\cite{kim2024let}. Subsequently, generative adversarial networks, such as VCA-GAN~\cite{kim2021lip}, were introduced to enhance realism of the synthesized audio.

Recently, diffusion models~\cite{Choi_2023_ICCV, choi2025v2sflow, yemini2023lipvoicer} have emerged as a powerful method for high-fidelity speech generation. However, while these generative models have significantly improved audio fidelity, ensuring precise textual content remains a persistent challenge. Some methods \cite{Choi_2023_ICCV, choi2025v2sflow} incorporate vision-guided speaker embeddings but lack explicit textual conditioning, thus failing to sufficiently mitigate the inherent one-to-many ambiguity. More recent work, such as LipVoicer \cite{yemini2023lipvoicer}, employs classifier guidance derived from lip-reading during the inference phase of the diffusion process to provide textual context. Critically, LipVoicer does not integrate text information during its training phase and relies on external classifier guidance for content correction. This approach necessitates an auxiliary classifier and, while achieving better textual alignment, often correlates with a loss of speech naturalness, which may partly stem from the external classifier's interference with the generative process. To overcome these limitations, it is essential to develop a robust, Classifier-Free Guidance (CFG) framework that effectively fuses textual and visual information from the training phase onward to ensure both high content accuracy and superior speech audio quality.

Our work builds on this direction by proposing a novel attention-based fusion module designed to explicitly align text and video features from the training phase onward. Furthermore, our proposed framework adopts a CFG strategy, eliminating the need for an auxiliary classifier and simplifying the overall architecture. This direct and persistent use of explicit textual conditioning, combined with our effective fusion strategy, effectively alleviates the inherent one-to-many ambiguity of the V2S task by constraining the mapping space. This approach aims to simultaneously improve the intelligibility, naturalness, and audio-visual synchronization of the generated speech.

\section{Methodology}
\label{sec:Methodology}

\begin{figure*}[t!]
\centering
\includegraphics[width=1.0\textwidth]{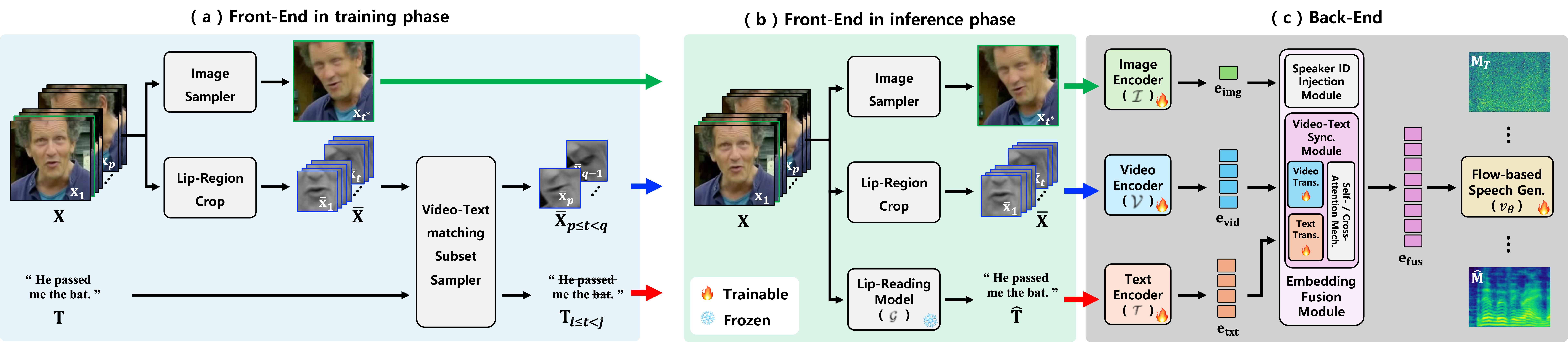} \vspace{1mm}
\caption{Overview of the proposed WYS framework: (a) training front-end using GT text $\mathbf{T}$, (b) inference front-end predicting text $\hat{\mathbf{T}}$ via a lip-reading model, and (c) shared back-end that encodes and fuses text, video, and speaker identity features to guide flow-based speech synthesis.}
\label{fig:architecture}
\end{figure*}

Our WYS framework is designed to generate high-fidelity mel-spectrograms that accurately preserve textual content while remaining tightly synchronized with the corresponding video sequences. By leveraging explicit textual conditioning and a novel embedding fusion module, WYS effectively alleviates the inherent one-to-many ambiguity of the V2S task, ensuring a more constrained mapping from visual-textual inputs to the target speech. As illustrated in Fig.~\ref{fig:architecture}, the overall architecture comprises four key components: \textbf{(1) a lip-reading module} for textual conditioning, \textbf{(2) feature extraction modules} for multimodal representation, \textbf{(3) an attention-based embedding fusion module} for cross-modal alignment, and \textbf{(4) a flow-matching-based model} for speech synthesis.

\subsection{Linguistic Information via Lip-Reading}
The WYS framework leverages textual sentences, $\mathbf{T}=(\mathbf{t}_1, \cdots, \mathbf{t}_{N})$, to provide explicit linguistic conditioning during both training and inference.

\paragraph{\textbf{Training}.} As shown in Fig.~\ref{fig:architecture}(a), the module is provided with GT text sentences, $\mathbf{T}_{i\leq t < j}$, corresponding to the sub-sampled video clips, $\mathbf{X}_{p\leq t <q}$. This pairing allows the model to learn a robust representation that integrates visual and textual cues, thereby narrowing down the mapping space toward the target speech $\mathbf{M}$. {Details of the sampling pipeline are provided in the supplementary material.}

\paragraph{\textbf{Inference.}} Since GT text is unavailable during inference, a pre-trained lip-reading model~\cite{ma2023auto}, denoted as $\mathcal{G}$, is employed to estimate a text sentence $\hat{\mathbf{T}}$ from the input silent videos $\mathbf{X}$. This process is formulated as $\hat{\mathbf{T}}=\mathcal{G}(\mathbf{X})$, as shown in Fig.~\ref{fig:architecture}(b).

\medskip\noindent
The resulting text $\tilde{\mathbf{T}}$—comprising either the GT text $\mathbf{T}_{i\leq t < j}$ during training or the predicted text $\hat{\mathbf{T}}$ during inference—is subsequently processed by a text encoder $\mathcal{T}$ to generate textual embeddings $\mathbf{e}_\text{txt}$. The detailed methodology for generating $\mathbf{e}_\text{txt}$ is further elaborated in the \textbf{Text Encoder} part of
. 3.2.

\subsection{Multimodal Representation via Feature Extraction}
The WYS framework receives a silent talking-face video as input, represented by a tensor $\mathbf{X} \in \mathbb{R}^{M \times H_\mathbf{x} \times W_\mathbf{x} \times C_\mathbf{x}}$, where $M$ is the number of frames, and $H_\mathbf{x}$, $W_\mathbf{x}$, and $C_\mathbf{x}$ are the height, width, and number of channels, respectively. From this raw input, three distinct streams are derived via preprocessing to provide comprehensive conditioning: \textbf{(i) a text sentence} for linguistic context, \textbf{(ii) a reference face frame} for speaker identity, and \textbf{(iii) a lip-region sequence} for temporal visual dynamics. A dedicated encoder subsequently processes each component to generate a representative feature embedding, as detailed in the following subsections.

\begin{figure}[t!]
    \centering
    \includegraphics[width=0.85\textwidth]{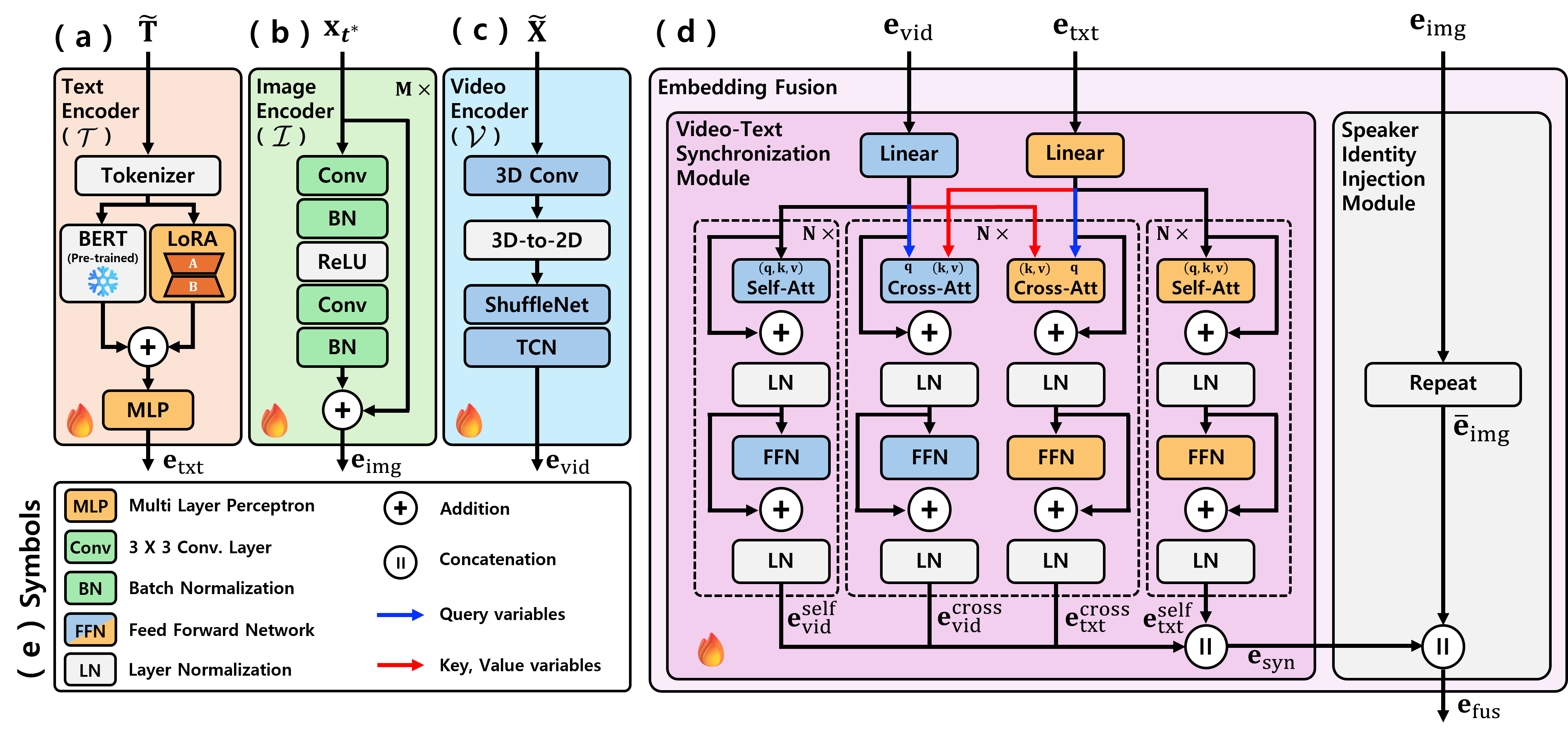} \vspace{2.5mm}
    \caption{Details of the back-end components: (a) Text Encoder, (b) Image Encoder for speaker identity, (c) Video Encoder for lip movements, (d) embedding fusion module that aligns video and text features via attention and injects speaker identity to produce $\mathbf{e}_\text{fus}$, and (e) legend.}
    \label{fig:architecture_details}
\end{figure}

\paragraph{\textbf{Text Encoder.}}
To extract robust linguistic representations from the input text, we utilize a pre-trained BERT~\cite{devlin2019bert} as the backbone of our text encoder. The input to the encoder is a text sentence $\tilde{\mathbf{T}}$, which represents the GT text $\mathbf{T}_{i \leq t < j}$ during training and the estimated text $\hat{\mathbf{T}}$ during inference. To facilitate efficient domain adaptation to the V2S task while preserving the pre-trained knowledge, we keep the BERT parameters $\mathbf{\theta}_\text{BERT}^*$ frozen and employ Low-Rank Adaptation (LoRA)~\cite{hu2022lora} for parameter-efficient fine-tuning. The architecture of the text encoder, denoted as $\mathcal{T}(\mathbf{\theta}_\text{BERT}^*, \mathbf{\theta}_\text{LoRA})$, is shown in Fig.~\ref{fig:architecture_details}(a). The process is formulated as:
\begin{equation}
    \mathbf{e}_\text{txt} = \mathcal{T}(\tilde{\mathbf{T}};~\mathbf{\theta}_\text{BERT}^*, \mathbf{\theta}_\text{LoRA}),
\label{eq:txt_enc}
\end{equation}
where $\mathbf{e}_\text{txt} \in \mathbb{R}^{L \times d_\text{txt}}$ is the resulting text embedding with sequence length $L$ and dimension $d_\text{txt}$. Here, $\mathbf{\theta}$ denotes the trainable weights, while $\mathbf{\theta}^*$ indicates the frozen parameters.

\paragraph{\textbf{Image Encoder.}}
To capture the speaker's identity, an image embedding $\mathbf{e}_\text{img}$ is extracted from a representative face frame $\mathbf{x}_{t^*} \in \mathbb{R}^{H_\mathbf{x} \times W_\mathbf{x} \times C_\mathbf{x}}$ sampled from the input video. As shown in Fig.~\ref{fig:architecture_details}(b), a ResNet18 \cite{he2016deep} is employed as the image encoder $\mathcal{I}(\theta_\text{img})$ to derive a global speaker representation. The process is formulated as:
\begin{equation}
    \mathbf{e}_\text{img} = \mathcal{I}({\mathbf{x}_{t^*}};~\mathbf{\theta}_\text{img}),
\label{eq:img_enc}
\end{equation}
where $\mathbf{e}_\text{img} \in \mathbb{R}^{1 \times d_\text{img}}$ is the speaker embedding, with $d_\text{img}$ representing the feature dimension of the image embedding.

\paragraph{\textbf{Video Encoder.}}
To capture the temporal dynamics of speech articulation, we first isolate the lip-region from the full face video $\mathbf{X}$ as follows:
\begin{equation}
    \hat{\mathbf{X}}=\text{LipCrop}(\mathbf{X}),
\label{eq:lip_crop}
\end{equation}
where $\hat{\mathbf{X}}$ is a localized lip-region video tensor, representing the primary visual stream for speech synthesis. To generate the video embedding $\mathbf{e}_\text{vid}$, we employ a video encoder $\mathcal{V}(\theta_\text{vid})$ based on a 3D CNN architecture \cite{gao2021visualvoice}, which is specifically designed to extract spatio-temporal features from facial movements. As shown in Fig.~\ref{fig:architecture_details}(c), this module consists of a 3D convolutional layer for local spatial-temporal processing, followed by a ShuffleNet~\cite{ma2018shufflenet} and a Temporal Convolutional Network (TCN)~\cite{lea2017temporal} to capture long-range temporal dependencies. The input $\tilde{\mathbf{X}}$ corresponds to a randomly sub-sampled clip $\hat{\mathbf{X}}_{i \leq t < j}$ during training and the full sequence $\hat{\mathbf{X}}$ during inference. The encoding process is formulated as:
\begin{equation}
    \mathbf{e}_\text{vid} = \mathcal{V}(\tilde{\mathbf{X}};~\mathbf{\theta}_\text{vid}),
\label{eq:vid_enc}
\end{equation}
where $\mathbf{e}_\text{vid} \in \mathbb{R}^{L \times d_\text{vid}}$ is the resulting video embedding, with $L$ and $d_\text{vid}$ denoting the sequence length and feature dimension, respectively.

\subsection{Cross-modal Alignment via Attention-based Embedding Fusion}
Once the textual context $\mathbf{e}_\text{txt}$, speaker identity $\mathbf{e}_\text{img}$, and visual dynamics $\mathbf{e}_\text{vid}$ are extracted, they are integrated into a unified representation to guide the speech synthesis process. The primary objective of the Attention-based Fusion Module is to explicitly align these multimodal features, thereby effectively mitigating the inherent one-to-many ambiguity by constraining the mapping space. As illustrated in Fig.~\ref{fig:architecture_details}(d), the module comprises two specialized components: \textbf{(i) a Video-Text Alignment module} for cross-modal synchronization and \textbf{(ii) a Speaker Identity Injection module} for global voice conditioning.
\vspace{-0.3cm}
\paragraph{\textbf{Video-Text Alignment Module.}}
The core of the fusion strategy is the video-text alignment module, which is engineered to produce a unified representation by synergizing video dynamics $\mathbf{e}_\text{vid}$ and text context $\mathbf{e}_\text{txt}$. This module employs a dual-stage attention mechanism to ensure that the linguistic constraints effectively resolve the visual ambiguities of lip movements.

Initially, to capture intra-modal dependencies, both embeddings are processed through $N$ self-attention blocks. Each block consists of a multi-head self-attention layer followed by a Feed-Forward Network (FFN), incorporating residual connections and layer normalization:
\begin{align}
    &\bar{\mathbf{e}}^{n}_{m} ~~~~= ~\text{LayerNorm}(\mathbf{e}_m^n + \text{Att}^\text{self}_{m}(\mathbf{e}_m^n)),  \nonumber \\
    &{\mathbf{e}}^{n+1}_{m} ~= ~\text{LayerNorm}(\bar{\mathbf{e}}^{n}_{m} + \text{FFN}(\bar{\mathbf{e}}^{n}_{m})),
\end{align}
where $\mathbf{e}_m^n \in \mathbb{R}^{L \times d_\text{self}}$ denotes the embedding at the $n$-th block for modality $m \in \{\text{vid}, \text{txt} \}$, and $d_\text{self}$ is the model dimension. This stage yields contextually enriched embeddings, ${\mathbf{e}}_\text{vid}^\text{self}$ and ${\mathbf{e}}_\text{txt}^\text{self}$.

Subsequently, to facilitate inter-modal interaction, $N$ cross-attention blocks are applied. In these blocks, one modality serves as the query while the other acts as the key and value, enabling bidirectional information exchange:
\begin{align}
    &\bar{\mathbf{e}}^{n}_{m} ~~~~= ~\text{LayerNorm}(\mathbf{e}_{m}^n + \text{Att}^\text{cross}_{m \to w}(\mathbf{e}_{m}^n, \mathbf{e}_{w}^n)),  \nonumber \\
    &{\mathbf{e}}^{n+1}_{m} ~= ~\text{LayerNorm}(\bar{\mathbf{e}}^{n}_{m} + \text{FFN}(\bar{\mathbf{e}}^{n}_{m})),
\end{align}
where $w$ represents the providing modality (i.e., if $m$ is video, $w$ is text, and vice versa). This results in two cross-modal embeddings, ${\mathbf{e}}_{\text{txt} \to \text{vid}}^\text{cross}$ and ${\mathbf{e}}_{\text{vid} \to \text{txt}}^\text{cross}$, which encode the mutual alignment between visual and linguistic tokens.

Finally, these four feature representations are concatenated to form the synchronized alignment embedding $\mathbf{e}_\text{syn}\in \mathbb{R}^{L \times d_\text{syn}}$:
\begin{equation}
    \mathbf{e}_\text{syn} = \left[~ {\mathbf{e}}_\text{vid}^\text{self} \parallel {\mathbf{e}}_\text{txt}^\text{self} \parallel {\mathbf{e}}_{\text{vid} \to \text{txt}}^\text{cross} \parallel {\mathbf{e}}_{\text{txt} \to \text{vid}}^\text{cross} ~\right],
\label{eq:fused_emb}
\end{equation}
where $\parallel$ is the concatenation and the dimension is $d_\text{syn} = 2\cdot d_\text{self} + 2\cdot d_\text{cross}$.
\vspace{-0.3cm}
\paragraph{\textbf{Speaker Identity Injection Module.}}
Since the speaker identity embedding $\mathbf{e}_\text{img} \in \mathbb{R}^{1 \times d_\text{img}}$ is extracted from a single reference frame, it possesses a sequence length of one. To integrate this static representation with the temporal synchronized embedding $\mathbf{e}_\text{syn} \in \mathbb{R}^{L \times d_\text{syn}}$, a sequence-matching process is required. We perform a temporal expansion of the identity embedding by broadcasting it across $L$ time steps, yielding $\bar{\mathbf{e}}_\text{img} \in \mathbb{R}^{L \times d_\text{img}}$. This expanded representation is then concatenated with $\mathbf{e}_\text{syn}$ to produce the final fused embedding $\mathbf{e}_\text{fus} \in \mathbb{R}^{L \times d_\text{fus}}$:
\begin{align}    &\bar{\mathbf{e}}_\text{img} ~= \text{Repeat}(\mathbf{e}_\text{img}, L), \\
    &\mathbf{e}_\text{fus} ~~= \left[ ~ \bar{\mathbf{e}}_\text{img} ~\parallel~ \mathbf{e}_\text{syn} ~\right],
\end{align}
where the fused embedding $\mathbf{e}_\text{fus}$ has a total dimension of $d_\text{fus} = d_\text{syn} + d_\text{img}$.

\subsection{Speech Synthesis via Conditional Flow Matching \\with Classifier-Free Guidance}
The final fused embedding $\mathbf{e}_\text{fus}$ serves as the conditional guidance for generating the target mel-spectrogram $\hat{\mathbf{M}}$. We adapt a CFM framework, which learns a vector field to transform a prior noise distribution into the target data distribution. We employ a DiffWave~\cite{kong2020diffwave} as the backbone to estimate the velocity field $v_\theta$.

To enhance the model's adherence to textual and visual constraints without requiring an external classifier, we incorporate CFG~\cite{ho2022classifier} during both training and inference. The CFM objective with CFG is defined as:
\begin{align}
    \mathcal{L}_{\text{CFM}} =~ &\mathbb{E}_{t, \mathbf{M}_0, \mathbf{e}_\text{fus}, \mathbf{x}_0,\mathbf{x}_1} \left[ \left\| v_\theta \left( {\mathbf{M}_t, t, \mathbf{e}_\text{fus}  } \right) -(\mathbf{x}_1-\mathbf{x}_0) \right\|^2 \right] \nonumber \\
    &~+ \mathbb{E}_{t, \mathbf{M}_0, \mathbf{x}_0,\mathbf{x}_1} \left[ \left\| v_\theta \left( {\mathbf{M}_t, t, \emptyset} \right) -(\mathbf{x}_1-\mathbf{x}_0)\right\|^2 \right],
\label{eq:CFG}
\end{align}
where $\mathbf{x}_1$ is the GT mel-spectrogram, $\mathbf{x}_0 \sim \mathcal{N}(0, \mathbf{I})$ is the Gaussian noise, and $\mathbf{M}_t = (1-t)~\mathbf{x}_0 + t~\mathbf{x}_1$ denotes the probability path at time $t \in [0, 1]$. During training, the condition $\mathbf{e}_\text{fus}$ is randomly replaced with a null token $\emptyset$, allowing the model to learn both conditional and unconditional distributions simultaneously.

During inference, we compute the guided velocity $\bar{v}_\theta$ by interpolating between the conditional and unconditional predictions:
\begin{equation}
\label{eq:CFG_infer}
    \bar{v}_\theta(\mathbf{M}_t ,t) = (1+\omega)\cdot v_\theta(\mathbf{M}_t,t,\textbf{e}_{\text{fus}})-\omega\cdot v_\theta(\mathbf{M}_t,t,\emptyset),
\end{equation}
where $\omega$ denotes the guidance scale. The mel-spectrogram is synthesized by solving the Ordinary Differential Equation (ODE) starting from $\textbf{M}_0 \sim \mathcal{N}(0, \mathbf{I})$:
\begin{equation}
\label{eq:sampling_infer}
    \textbf{M}_{t-\Delta t} = \textbf{M}_{t} - \Delta t \cdot \bar{v}_\theta(\textbf{M}_t,t).
\end{equation}

The final generated mel-spectrogram $\hat{\textbf{M}} = \textbf{M}_1$ is then converted into the speech waveform $\hat{\textbf{Y}}$ using a pre-trained HiFi-GAN vocoder~\cite{kong2020hifi}:
\begin{equation}
    \hat{\mathbf{Y}} = \text{Vocoder}(\hat{\mathbf{M}}).
\label{eq:vocoder}
\end{equation}

\section{Experimental Setup}
\label{sec:Experiments_Setup}

\subsection{Datasets}\label{subsec:Dataset}
We evaluate the proposed framework on two large-scale audio-visual benchmarks: \textbf{LRS2}~\cite{afouras2018a}, sourced from BBC broadcasts (144K clips, 224 hours, $>$13K vocabulary), and \textbf{LRS3}~\cite{afouras2018b}, derived from TED talks (151K clips, 475 hours, $>$40K vocabulary) with significantly higher variability in speaker demographics, accents, and recording conditions. All videos are processed at 25 frame per second (fps) with audio sampled at 16 kHz. {Details of the preprocessing pipeline and dataset splits are provided in the supplementary material.}

\subsection{Implementation Details}
The textual and visual  embedding dimensions are set to $d_\text{txt} = d_\text{vid} = 512$, while the image encoder produces $d_\text{img} = 128$. The fusion module uses $\textbf{N} = 6$, attention blocks with $h = 8$ heads, and hidden dimensions $d_\text{self} = d_\text{cross} = 512$. The model was trained on a single NVIDIA RTX PRO 6000 GPU using the Adam optimizer ($\text{lr} = 2\times10^{-4}$, batch size 64) for 1,000,000 iterations with a null-conditioning probability of 0.1 for CFG. During inference, we use a first-order Euler ODE solver with $T = 100$ steps and guidance scale $\omega = 2.0$.

\subsection{Evaluation Metrics}
To provide a comprehensive evaluation, we assess the proposed framework using both \textbf{objective quantitative metrics} and \textbf{subjective qualitative analysis}.

\paragraph{\textbf{Quantitative Metrics.}}
We evaluate the performance of the WYS framework across three primary dimensions: content accuracy, audio-visual synchronization, and speech quality. To assess {content accuracy}, we compute the \textbf{Word Error Rate (WER)} by transcribing the generated audio via a pre-trained Automatic Speech Recognition (ASR) model~\cite{ma2023auto} and comparing the transcript against the GT text. {Audio-visual synchronization} is quantified using the \textbf{LSE-C} (Confidence) and \textbf{LSE-D} (Distance) metrics from a pre-trained SyncNet~\cite{chung2016}, which measure the temporal alignment between the input lip movements and the synthesized speech. Finally, we evaluate the {perceptual quality and intelligibility} of the audio signals using \textbf{STOI-Net}~\cite{zezario2020} and \textbf{DNSMOS}~\cite{chandan2022}, providing an objective measure of the clarity and human-likeness of the generated utterances.

\paragraph{\textbf{Qualitative Analysis.}}
For qualitative assessment, we conducted human listening tests via Amazon Mechanical Turk. A total of 100 participants evaluated 10 randomly selected samples from each dataset across the proposed and four baseline models. Raters provided scores on a 5-point Mean Opinion Score (MOS) scale based on five criteria: (1) \textbf{Audio Quality} (overall acoustic fidelity), (2) \textbf{Content Accuracy} (correspondence with the intended text), (3) \textbf{Intelligibility} (clarity of the spoken words), (4) \textbf{Lip-sync} (audio-visual synchronization), and (5) \textbf{Naturalness} (the human-likeness of the synthesized voice). {The specific questionnaire and survey interface utilized for the Amazon Mechanical Turk evaluations are provided in the supplementary material.}

\section{Experimental Results}
\label{sec:Experiments_Result}
To validate the effectiveness of the proposed WYS framework, we conduct extensive comparative experiments against various baselines encompassing a broad spectrum of generative paradigms in the video-to-speech field: \textbf{VCA-GAN}~\cite{kim2021lip} (GAN-based model), \textbf{SVTS}~\cite{mira2022svts} (Conformer-based model), and \textbf{IntelligibleL2S}~\cite{choi2023intelligible} (Speech Unit-based model). Furthermore, we compare our work with recent high-performance generative models: \textbf{DiffV2S}~\cite{Choi_2023_ICCV} (Diffusion model without text guidance), \textbf{LipVoicer}~\cite{yemini2023lipvoicer} (Diffusion model with inference-time text guidance), \textbf{V2SFlow}~\cite{choi2025v2sflow} (Rectified flow matching model)\footnote{We use V2SFlow-V (video-derived embeddings) for fair comparison, as reference audio is unavailable in the silent V2S setting.}, \textbf{FTV}~\cite{kim2025faces} (Hierachical representation model), and \textbf{AlignDiT}~\cite{choi2025aligndit} (Diffusion transformer based model).

Through these comparisons, we demonstrate the superiority of our integrated textual conditioning and CFG strategy in producing intelligible and natural speech. {Additional visualizations of the generated mel-spectrograms are provided in the supplementary material for further qualitative inspection.}

\subsection{Main Results}
\paragraph{\textbf{Quantitative Results.}}
As summarized in Tab.~\ref{tab:quan_result}, the WYS framework demonstrates highly competitive performance across all evaluation dimensions. A critical observation arises from the comparison with LipVoicer \cite{yemini2023lipvoicer}. While LipVoicer achieves slightly lower WER scores (18.8991\% vs. 20.1321\% on LRS2; 22.2423\% vs. 23.7349\% on LRS3), the proposed framework consistently outperforms LipVoicer across all other metrics, including synchronization (LSE-C/D) and perceptual quality (STOI-Net, DNSMOS).

This strategic trade-off highlights a fundamental difference in methodology. LipVoicer’s marginal lead in WER can be attributed to its reliance on external classifier guidance during inference, which acts as a rigid content-correction tool. While this mechanism forces the output to align with the text, it appears to do so at the expense of audio-visual integrity and naturalness. Such discriminative intervention during the generative process likely disrupts the synchronization between lip movements and speech (as reflected in poorer LSE-C/D scores) and degrades the overall acoustic fidelity (indicated by lower DNSMOS results).

In contrast, our results underscore the effectiveness of an integrated training-time fusion strategy. By embedding textual constraints directly into the model's training phase, the WYS framework achieves a refined balance between linguistic fidelity and high-fidelity synthesis. These findings demonstrate that our approach provides a more robust solution for V2S, effectively constraining the ill-posed mapping space between visual cues and acoustic features without compromising the inherent naturalness of the synthesized speech.

\begin{table}[t!]
\centering
\resizebox{\textwidth}{!}{%
\begin{tabular}{l ccccc ccccc}
\toprule
\multirow{2}{*}{\textbf{Method}}
& \multicolumn{5}{c}{\textbf{LRS2-BBC}}
& \multicolumn{5}{c}{\textbf{LRS3-TED}} \\
\cmidrule(lr){2-6} \cmidrule(lr){7-11}
& \textbf{WER} [\%] $\downarrow$
& \textbf{LSE-C} $\uparrow$
& \textbf{LSE-D} $\downarrow$
& \textbf{STOI-Net} $\uparrow$
& \textbf{DNSMOS} $\uparrow$
& \textbf{WER} [\%] $\downarrow$
& \textbf{LSE-C} $\uparrow$
& \textbf{LSE-D} $\downarrow$
& \textbf{STOI-Net} $\uparrow$
& \textbf{DNSMOS} $\uparrow$ \\
\midrule

Ground Truth
& 1.4773 & 6.9788 & 7.1891 & 0.9063 & 3.1384
& 1.0096 & 7.3393 & 6.9002 & 0.9307 & 3.2988 \\

\midrule

VCA-GAN~
& 100.7812 & 2.3960 & 11.7631 & 0.5107 & 2.2573
& 90.6827 & 4.5511 & 9.1256 & 0.6325 & 2.2672 \\

SVTS~
& -- & -- & -- & -- & --
& 75.6552 & 6.2841 & 7.9446 & 0.7070 & 2.4203 \\

DiffV2S~
& 50.7801 & 6.4320 & 7.6540 & 0.8923 & 2.9230
& 38.0897 & 6.2841 & 7.9403 & 0.9214 & 3.2169 \\

IntelligibleL2S~
& 44.2286 & 7.1280 & \underline{7.0210} & 0.8588 & 2.7063
& 50.0105 & 6.7813 & 7.3754 & 0.8838 & 2.8678 \\

LipVoicer~
& \textbf{18.8991} & 6.5512 & 7.8622 & 0.9034 & 3.0352
& \textbf{22.2423} & 5.7035 & 8.4518 & 0.9207 & 3.1846 \\

V2SFlow-V~
& 36.1311 & \underline{7.1897} & 7.2642 & \textbf{0.9225} & \textbf{3.1131}
& 29.6079 & 7.0084 & 7.4493 & 0.9328 & \textbf{3.2851} \\

FTV~
& -- & -- & -- & -- & --
& 29.1370 & \textbf{7.0996} & \underline{7.2633} & \underline{0.9348} & 3.1685 \\

AlignDiT~
& -- & -- & -- & -- & --
& 29.6671  & 6.9625 & 7.3483 & 0.9331 & 3.1959 \\

\rowcolor{gray!15}
\textbf{Ours (WYS)}
& \underline{20.1321} & \textbf{7.5032} & \textbf{6.8118} & \underline{0.9144} & \underline{3.0927}
& \underline{23.7349} & \underline{7.0148} & \textbf{7.2586} & \textbf{0.9373} & \underline{3.2414} \\

\bottomrule
\end{tabular}
}\vspace{0.3cm}
\caption{Quantitative results on LRS2-BBC and LRS3-TED. \textbf{Best} and \underline{second-best} scores are highlighted. $\downarrow$/$\uparrow$ denote lower/higher is better.}
\label{tab:quan_result}
\end{table}

\begin{table}[t!]
\centering
\resizebox{\textwidth}{!}{%
\begin{tabular}{l ccccc ccccc}
\toprule
\multirow{2}{*}{\textbf{Method}}
& \multicolumn{5}{c}{\textbf{LRS2-BBC}}
& \multicolumn{5}{c}{\textbf{LRS3-TED}} \\
\cmidrule(lr){2-6} \cmidrule(lr){7-11}
& \textbf{Qual.} $\uparrow$
& \textbf{Align.} $\uparrow$
& \textbf{Intel.} $\uparrow$
& \textbf{Sync.} $\uparrow$
& \textbf{Natural.} $\uparrow$
& \textbf{Qual.} $\uparrow$
& \textbf{Align.} $\uparrow$
& \textbf{Intel.} $\uparrow$
& \textbf{Sync.} $\uparrow$
& \textbf{Natural.} $\uparrow$ \\
\midrule

Ground Truth
    & 4.26 $\pm$ 0.15 & 4.36 $\pm$ 0.11
    & 4.31 $\pm$ 0.13 & 4.26 $\pm$ 0.14
    & 4.27 $\pm$ 0.10
    & 4.25 $\pm$ 0.12 & 4.33 $\pm$ 0.10
    & 4.23 $\pm$ 0.11 & 4.23 $\pm$ 0.11
    & 4.28 $\pm$ 0.01 \\

\midrule

IntelligibleL2S
    & 2.77 $\pm$ 0.23 & 3.06 $\pm$ 0.25
    & 3.00 $\pm$ 0.24 & 3.36 $\pm$ 0.22
    & 2.87 $\pm$ 0.24
    & 2.59 $\pm$ 0.23 & 3.02 $\pm$ 0.26
    & 2.85 $\pm$ 0.24 & 3.08 $\pm$ 0.23
    & 2.82 $\pm$ 0.24 \\

DiffV2S
    & 3.24 $\pm$ 0.21 & 3.13 $\pm$ 0.24
    & 3.24 $\pm$ 0.23 & 3.51 $\pm$ 0.19
    & 3.27 $\pm$ 0.21
    & 3.56 $\pm$ 0.20 & 3.55 $\pm$ 0.23
    & 3.48 $\pm$ 0.23 & 3.70 $\pm$ 0.20
    & 3.64 $\pm$ 0.20 \\

LipVoicer
    & 3.49 $\pm$ 0.21 & \underline{3.83 $\pm$ 0.16}
    & 3.64 $\pm$ 0.17 & 3.73 $\pm$ 0.18
    & 3.56 $\pm$ 0.19
    & 3.71 $\pm$ 0.21 & \underline{3.87 $\pm$ 0.19}
    & 3.77 $\pm$ 0.21 & \underline{3.90 $\pm$ 0.18 }
    & 3.65 $\pm$ 0.20 \\

V2SFlow-V
    & \underline{3.73 $\pm$ 0.19} & 3.42 $\pm$ 0.24
    & \underline{3.69 $\pm$ 0.18} & \underline{3.83 $\pm$ 0.19}
    & \underline{3.85 $\pm$ 0.19}
    & \underline{3.86 $\pm$ 0.22} & 3.77 $\pm$ 0.22
    & \underline{3.78 $\pm$ 0.22} & 3.83 $\pm$ 0.23
    & \underline{3.77 $\pm$ 0.20} \\

\rowcolor{gray!15}
\textbf{Ours (WYS)}
    & \textbf{3.92 $\pm$ 0.19} & \textbf{3.99 $\pm$ 0.19}
    & \textbf{3.97 $\pm$ 0.18} & \textbf{4.03 $\pm$ 0.18}
    & \textbf{4.00 $\pm$ 0.10}
    & \textbf{3.93 $\pm$ 0.18} & \textbf{4.03 $\pm$ 0.18}
    & \textbf{4.02 $\pm$ 0.19} & \textbf{4.01 $\pm$ 0.17}
    & \textbf{3.93 $\pm$ 0.18} \\

\bottomrule
\end{tabular}
}\vspace{0.3cm}
\caption{Subjective evaluation (MOS) on LRS2-BBC and LRS3-TED. \textbf{Best} and \underline{second-best} scores are highlighted. $\uparrow$ denotes higher is better.}
\label{tab:qual_result}
\end{table}

\begin{figure}[t!]
    \centering
    \includegraphics[width=0.7\textwidth]{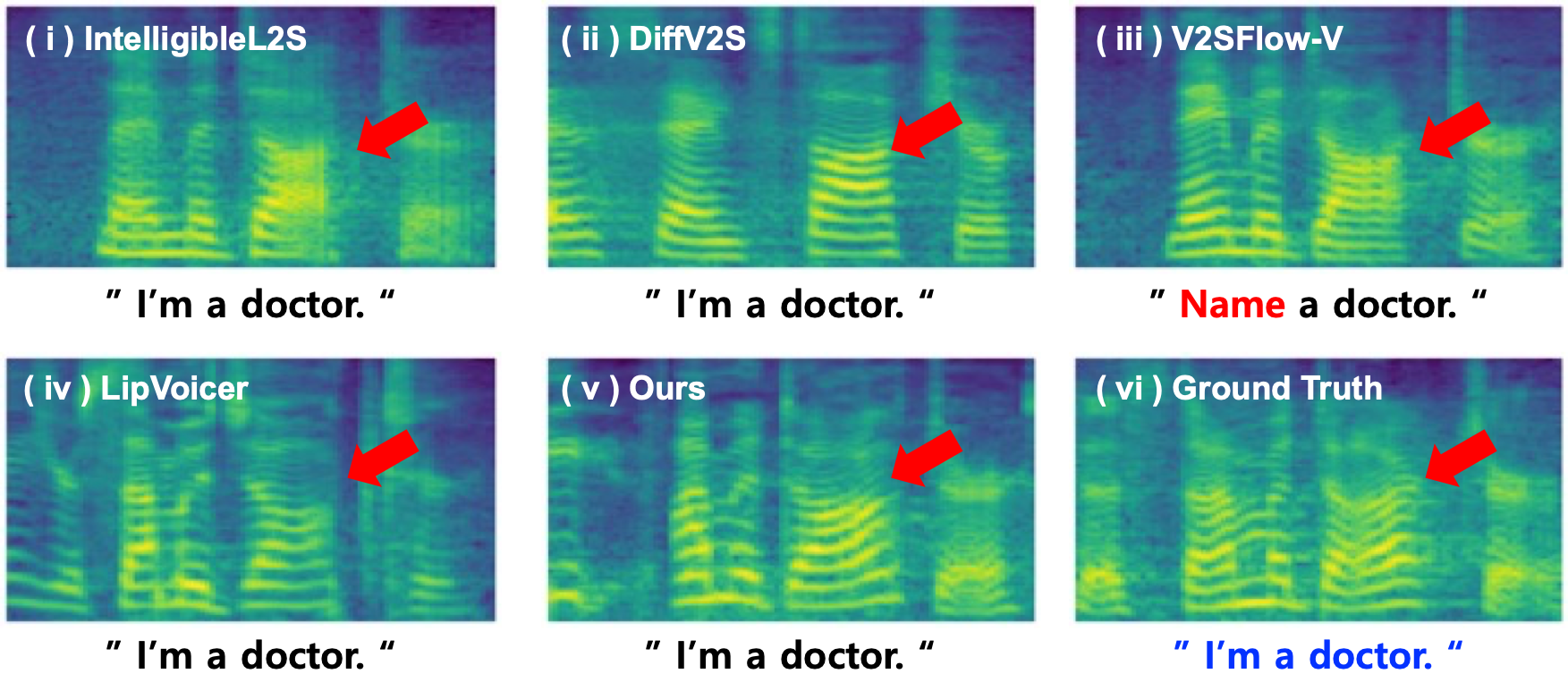} \vspace{0.2cm}
    \caption{Qualitative comparison of mel-spectrograms on LRS2. The text below each spectrogram is predicted by an ASR model, where \textcolor{blue}{GT text} is shown in blue and \textcolor{red}{incorrectly predicted words} are highlighted in red.}
    \label{fig:mel_one_sample}
\end{figure}

\paragraph{\textbf{Qualitative Results.}}
The results of the subjective evaluation are summarized in Tab.~\ref{tab:qual_result}, including the 95\% confidence intervals for each score.
In contrast to the subtle differences observed in some quantitative results, the proposed WYS framework significantly outperforms all baseline methods across every perceptual criterion on both datasets.

This perceptual superiority is further substantiated by the visual analysis of synthesized mel-spectrograms in Fig.~\ref{fig:mel_one_sample}. Our model in Fig.~\ref{fig:mel_one_sample}(v) not only demonstrates high acoustic structural fidelity compared to the GT text (see Fig.~\ref{fig:mel_one_sample}(vi)) but also maintains precise linguistic integrity. In contrast, V2SFlow-V (see Fig.~\ref{fig:mel_one_sample}(iii)) frequently fails to recover the correct textual content, resulting in phonetic mismatches. While other baselines, such as IntelligibleL2S and LipVoicer, achieve better content alignment, they exhibit noticeable structural distortions and a lack of fine-grained acoustic details as shown in Fig.~\ref{fig:mel_one_sample}(i, ii).

These findings indicate that the WYS framework generates intelligible and realistic speech while maintaining alignment with the intended textual context. By bridging linguistic cues and visual dynamics, our model achieves improved naturalness and synchronization compared with baseline methods.

\begin{table}[t!]
\centering
\resizebox{0.8\textwidth}{!}{
\begin{tabular}{cccccc}
  \toprule
    \multirow{2}{*}{\shortstack{\textbf{Attention}\\\textbf{Mechanisms}}} &
    \multicolumn{1}{c}{~~\textbf{Textual Acc.}~} &
    \multicolumn{2}{c}{~\textbf{A-V Sync.}~} &
    \multicolumn{2}{c}{~\textbf{Audio Qual.}~~} \\
    \cmidrule(lr){2-2} \cmidrule(lr){3-4} \cmidrule(lr){5-6}
    & ~\textbf{WER} [\%] $\downarrow$~
    & ~\textbf{LSE-C} $\uparrow$~
    & ~\textbf{LSE-D} $\downarrow$~
    & ~\textbf{STOI-Net} $\uparrow$~
    & ~\textbf{DNSMOS} $\uparrow$~~ \\
    \cmidrule(lr){1-6}
  $\mathbf{e}_\text{txt} \parallel \mathbf{e}_\text{vid} $
  & 50.6510
  & \textbf{7.6259}
  & \textbf{6.7147}
  & 0.9093
  & 3.0811 \\
  $\mathbf{e}_\text{txt}^\text{self} \parallel \mathbf{e}_\text{vid}^\text{self}$
  & 52.3977
  & \underline{7.6127}
  & \underline{6.7217}
  & 0.9128
  & 3.0581 \\
  $\mathbf{e}_{\text{txt} \to \text{vid}}^\text{cross}$
  & 80.8254
  & 5.5675
  & 8.6043
  & 0.8982
  & 3.0389 \\
  $\mathbf{e}_{\text{vid} \to \text{txt}}^\text{cross}$
  & 21.0629
  & 7.5863
  & 6.7593
  & 0.9101
  & 3.0888 \\
  $\mathbf{e}_{\text{txt} \to \text{vid}}^\text{cross} \parallel \mathbf{e}_{\text{vid} \to \text{txt}}^\text{cross}$
  & \underline{20.8882}
  & 7.5219
  & 6.7984
  & \underline{0.9113}
  & \underline{3.0902} \\
  \rowcolor{gray!15} \textbf{Ours} ($\mathbf{e}_\text{syn}$)
  &  \textbf{20.1321}
  & 7.5032
  & 6.8118
  & \textbf{0.9144}
  & \textbf{3.0927} \\
  \bottomrule
\end{tabular}
}
\vspace{0.4cm}
\caption{Ablation study on the attention mechanism (LRS2). $\mathbf{e}_{\text{m}}^\text{self}$: Self-attention embedding; $\mathbf{e}_{m \to w}^\text{cross}$: Cross-attention embedding w/ $m$ as query; $\mathbf{e}_\text{syn}$: Our fused embedding.}
\label{tab:att_abl}
\end{table}

\subsection{Ablation Studies}
To evaluate our design choices and the individual contributions of each component, we conduct four ablation studies on the LRS2 dataset. Specifically, we investigate the impact of the \textbf{attention-based fusion}, the \textbf{speaker identity module}, the \textbf{guidance scale factor}, and the \textbf{sampling step count}. These experiments justify our architectural configuration and demonstrate the framework's robustness in balancing linguistic precision with acoustic naturalness.

\paragraph{\textbf{Effect of Attention Mechanism.}}
To verify the synergy between multimodal components, we compared our proposed fusion embedding $\mathbf{e}_\text{syn}$ with simpler variants in Tab.~\ref{tab:att_abl}. A baseline using only feature concatenation ($\mathbf{e}_\text{txt} \parallel \mathbf{e}_\text{vid}$) yielded a high WER of 50.6510\%, proving that raw merging is insufficient for content recovery despite maintaining decent audio-visual synchronization. Similarly, incorporating only self-attention ($\mathbf{e}_\text{txt}^\text{self} \parallel \mathbf{e}_\text{vid}^\text{self}$) independently refines each modality but still fails to integrate correct context, resulting in a high WER of 52.3977\%.

The most critical gain in content accuracy stems from the direction of cross-attention. Using video features as the query to attend to text ($\mathbf{e}_{\text{vid} \to \text{txt}}^{\text{cross}}$) dramatically lowers the WER to 21.0629\%, confirming that the temporal structure of the video must drive the extraction of linguistic content. In contrast, using text as the query ($\mathbf{e}_{\text{txt} \to \text{vid}}^{\text{cross}}$) fails to ground the content in the visual timeline, leading to a WER of 80.8254\%. Furthermore, while incorporating both cross-attention directions ($\mathbf{e}_{\text{txt} \to \text{vid}}^{\text{cross}} \parallel \mathbf{e}_{\text{vid} \to \text{txt}}^{\text{cross}}$) causes a slight degradation in LSE-C/D compared to the single $(\text{vid} \to \text{txt})$ model, it achieves a further WER reduction to 20.8882\%. This indicates that although the $(\text{txt} \to \text{vid})$ direction is less effective in isolation, it provides a valuable synergistic effect for content accuracy when integrated.

Our final fusion strategy, $\mathbf{e}_{\text{syn}}$, achieves the optimal balance by combining contextually refined self-attention features with dual cross-attention. This holistic approach yields the lowest overall WER (20.1321\%) and maximizes audio quality (STOI-Net: 0.9144, DNSMOS: 3.0927) while maintaining highly competitive synchronization (LSE-C: 7.5032, LSE-D: 6.8118), effectively bridging the gap between linguistic precision and acoustic naturalness.

\begin{table}[t!]

\centering
\resizebox{0.80\textwidth}{!}{
\renewcommand{\arraystretch}{1.2}
\begin{tabular}{l cccccc}
\toprule
\textbf{Configuration} & \textbf{WER} [\%] $\downarrow$ & \textbf{LSE-C} $\uparrow$ & \textbf{LSE-D} $\downarrow$ & \textbf{STOI-Net} $\uparrow$ & \textbf{DNSMOS} $\uparrow$ & \textbf{spkSIM} $\uparrow$ \\
\midrule
Ours ($\mathbf{e}_\text{img} = \mathbf{0}$) & 20.5793 & 7.5023 & 6.8248 & 0.9130 & 3.0783 & 0.6198 \\
\rowcolor{gray!15}
\textbf{Ours (WYS)} & \textbf{20.1321} & \textbf{7.5032} & \textbf{6.8118} & \textbf{0.9144} & \textbf{3.0927} & \textbf{0.6823} \\
\midrule
\textbf{Relative Change [\%]} & {2.2213} & {0.0119} & {0.0498} & {0.1533} & {0.4677} & \textbf{{10.0839}} \\
\bottomrule
\end{tabular}
}
\vspace{0.4cm}
\caption{Ablation study on the speaker identity module (LRS2).}\label{tab:spk_id_zero}
\end{table}

\begin{table}[t!]

\centering
\resizebox{0.6\textwidth}{!}{
\begin{tabular}{c c c c c c}
  \toprule
  \multirow{2}{*}{~~$\omega$~} & \multicolumn{1}{c}{~\textbf{Textual Acc.}~} & \multicolumn{2}{c}{~\textbf{A-V Sync.}~} & \multicolumn{2}{c}{~\textbf{Audio Qual.}~} \\
    \cmidrule(lr){2-2} \cmidrule(lr){3-4} \cmidrule(lr){5-6}
    & ~\textbf{WER} [\%] $\downarrow$~
    & ~\textbf{LSE-C} $\uparrow$~
    & ~\textbf{LSE-D} $\downarrow$~
    & ~\textbf{STOI-Net} $\uparrow$~
    & ~\textbf{DNSMOS} $\uparrow$~~ \\
    \cmidrule(lr){1-6}
  -1
  & 103.8737
  & 2.3689
  & 11.7148
  & 0.8890
  & 2.8528 \\
  0.0
  & 28.0110
  & 6.4583
  & 7.6201
  & 0.8946
  & 2.9162 \\
  1.0
  & 21.0289
  & \underline{7.4421}
  & \underline{6.8604}
  & 0.9111
  & \underline{3.0711} \\
  \rowcolor{gray!15} \textbf{2.0}
  & \textbf{20.1321}
  & \textbf{7.5032}
  & \textbf{6.8118}
  & \textbf{0.9144}
  & \textbf{3.0927} \\
  3.0
  & \underline{20.5699}
  & 7.4031
  & 6.8713
  & \underline{0.9120}
  & 3.0720 \\
  \bottomrule
\end{tabular}
}
\vspace{0.4cm}
\caption{Ablation study on the guidance scale factor $\omega$ (LRS2).} \label{tab:gui_abl}
\end{table}

\vspace{-0.5cm}
\paragraph{\textbf{Effect of Speaker Identity Module.}}
The contribution of the identity encoder is evaluated by zeroing out the image input ($\mathbf{e}_{\text{img}}=\mathbf{0}$), with the results summarized in Tab.~\ref{tab:spk_id_zero}. A quantitative analysis reveals a striking disparity in how the removal of identity embeddings affects the model. While core performance metrics exhibit only {marginal fluctuations}, the speaker similarity score ({spkSIM}) undergoes a {significant drop of 10.0839\%}. This phenomenon confirms that the speaker identity module specifically governs the preservation of individual voice characteristics without interfering with linguistic intelligibility or audio-visual synchronization. These findings suggest {a clear functional separation between the identity injection and cross-modal fusion modules}, demonstrating that each component focuses on its designated role.

\paragraph{\textbf{Effect of Guidance Scale Factor.}}
We investigated the impact of the CFG scale factor $\omega$, which modulates the balance between conditional and unconditional predictions. As shown in Tab.~\ref{tab:gui_abl}, the guidance mechanism is fundamental to the model's performance. Disabling conditioning entirely ($\omega = -1$) leads to a catastrophic collapse in all metrics, with the WER exceeding 100\%. Performance consistently improves as $\omega$ approaches 2.0, where the model reaches its {optimal peak} across all evaluation dimensions, including Textual accuracy, A-V synchronization, and Audio quality. Beyond this point, as seen with $\omega = 3.0$, the scores begin to marginally decline. Accordingly, we adopt $\omega = 2.0$ as the default for all experiments to ensure the best balance between linguistic precision and acoustic quality.

\begin{table}[t!]

\centering
\resizebox{0.8\textwidth}{!}{
\begin{tabular}{c c c c c c c}
  \toprule
    \multirow{2}{*}{~~\textbf{Steps}~} & \multicolumn{1}{c}{~\textbf{Textual Acc.}~} & \multicolumn{2}{c}{~\textbf{A-V Sync.}~} & \multicolumn{2}{c}{~\textbf{Audio Qual.}~} & \multicolumn{1}{c}{~\textbf{Sampling Time}~~} \\
    \cmidrule(lr){2-2} \cmidrule(lr){3-4} \cmidrule(lr){5-6} \cmidrule(lr){7-7}
    & ~\textbf{WER} [\%] $\downarrow$~
    & ~\textbf{LSE-C} $\uparrow$~
    & ~\textbf{LSE-D} $\downarrow$~
    & ~\textbf{STOI-Net} $\uparrow$~
    & ~\textbf{DNSMOS} $\uparrow$~
    & ~\textbf{(sec/sample)}~~ \\
    \cmidrule(lr){1-7}
  10
  & \underline{20.5293}
  & \textbf{7.5808}
  & \textbf{6.7572}
  & \underline{0.9130}
  & \underline{3.0783}
  & \textbf{0.7} \\
  \rowcolor{gray!15} \textbf{100}
  & \textbf{20.1321}
  & \underline{7.5032}
  & \underline{6.8118}
  & \textbf{0.9144}
  & \textbf{3.0927}
  & \underline{8.8} \\
  1000
  & 20.6555
  & 7.4850
  & 6.8194
  & 0.9127
  & 3.0468
  & 84.2 \\
  \bottomrule
\end{tabular}
}
\vspace{0.3cm}\caption{Ablation study on the number of sampling steps (LRS2).} \label{tab:step_abl}
\end{table}

\paragraph{\textbf{Effect of Sampling Step.}}
We analyzed the trade-off between inference efficiency and synthesis quality by varying the number of sampling steps $T$, as summarized in Tab.~\ref{tab:step_abl}. While $T = 100$ provides the most refined balance across all metrics, extending the process to $T = 1000$ yields negligible improvements while significantly increasing latency to 84.2 sec per sample. Notably, a minimal configuration of $T = 10$ maintains surprisingly robust performance, drastically accelerating inference to 0.7 sec per sample. These results demonstrate the high sampling efficiency of our rectified flow matching framework, suggesting that a low-step configuration is highly effective for real-time applications where low latency is prioritized without substantial loss in speech quality.

\begin{table}[t!]
\centering
\small
\renewcommand{\arraystretch}{0.8}
\setlength{\tabcolsep}{8.0pt}
\resizebox{0.9\textwidth}{!}{
\begin{tabular}{l ccccc}
  \toprule
    \multirow{2}{*}{\shortstack{\textbf{Method}}} &
    \multicolumn{1}{c}{~~\textbf{Textual Acc.}~} &
    \multicolumn{2}{c}{~\textbf{A-V Sync.}~} &
    \multicolumn{2}{c}{~\textbf{Audio Qual.}~~} \\
    \cmidrule(lr){2-2} \cmidrule(lr){3-4} \cmidrule(lr){5-6}
    & ~\textbf{WER} [\%] $\downarrow$~
    & ~\textbf{LSE-C} $\uparrow$~
    & ~\textbf{LSE-D} $\downarrow$~
    & ~\textbf{STOI-Net} $\uparrow$~
    & ~\textbf{DNSMOS} $\uparrow$~~ \\
    \cmidrule(lr){1-6}
  \multicolumn{6}{l}{\emph{(a) LRS2-BBC: inference-text source}}\\
  \textbf{Ours} (w/ GT Text)
  & \textbf{9.9784}
  & 7.4889
  & 6.8198
  & 0.9136
  & 3.0901 \\
  \rowcolor{gray!15} \textbf{Ours} (w/ Pred. Text)
  & 20.1321
  & \textbf{7.5032}
  & \textbf{6.8118}
  & \textbf{0.9144}
  & \textbf{3.0927} \\
  \midrule
  \multicolumn{6}{l}{\emph{(b) LRS2-BBC: training-text source}}\\
  \textbf{Ours} (w/ Pred.\ Text) & 40.8559 & \textbf{7.7031} & \textbf{6.6299} & \textbf{0.9173} & 3.0908 \\
  \rowcolor{gray!15}\textbf{Ours} (w/ GT Text) & \textbf{20.1321} & 7.5032 & 6.8118 & 0.9144 & \textbf{3.0927} \\
    \midrule
  \multicolumn{6}{l}{\emph{(c) LRS2-BBC: Lipreading+TTS pipeline}}\\
  XTTS-V2 & 31.4021 & 3.1225 & 11.3694 & 0.8060 & \textbf{3.3773} \\
  CosyVoice & \underline{29.4339} & \underline{3.7121} & \underline{10.9601} & \underline{0.9093} & \underline{3.2531} \\
  \rowcolor{gray!15}\textbf{Ours} & \textbf{20.1321} & \textbf{7.5032} & \textbf{6.8118} & \textbf{0.9144} & 3.0927 \\
  \midrule
  \multicolumn{6}{l}{\emph{(d) LRS3\(\to\)LRS2: cross-dataset}}\\
  LipVoicer & \underline{39.9212} & 5.6128 & 8.6038 & 0.9142 & 3.0505 \\
  V2SFlow-V & 46.1765 & \underline{7.2292} & \underline{7.2685} & \textbf{0.9356} & \underline{3.0865} \\
  \rowcolor{gray!15}\textbf{Ours} & \textbf{34.8539} & \textbf{7.3421} & \textbf{7.0123} & \underline{0.9299} & \textbf{3.1383} \\
  \bottomrule
\end{tabular}}
\vspace{0.4cm}\caption{Results of additional experiments. (a) inference-text source on LRS2, (b) training-text source on LRS2, (c) Lipreading+TTS on LRS2, (d) cross-dataset LRS3\(\to\)LRS2.} %
\label{tab:additional_exp}
\end{table}
\subsection{Additional Experiments}

\paragraph{\textbf{Performance Upper Bound and Linguistic Sensitivity Analysis.}}
We established the performance upper bound of the WYS framework by comparing predicted lip-read text against GT text conditioning. As summarized in Tab.~\ref{tab:additional_exp} (a), providing GT text substantially reduces the WER from 20.1321\% to 9.9784\%, nearly doubling the speech intelligibility. Essentially, as lip-reading technology matures, the WYS framework is perfectly poised to ``scale up'' its performance, directly translating future linguistic accuracy into superior acoustic synthesis.

\paragraph{\textbf{Training with Predicted Text.}}
In Tab.~\ref{tab:additional_exp}(b), training with GT text halves the WER (20.1 vs. 40.9),while the other four metrics remain comparable. This suggests that noisy predicted text mainly degrades linguistic supervision rather than A-V synchronization or audio quality. Training with unreliable pseudo-text can encourage the model to ignore or distrust textual conditioning, degrading its ability to generate accurate speech content. We therefore use GT text during training and predicted text only at inference.

\paragraph{\textbf{Lipreading + TTS baseline.}}
We provide the same predicted text to strong zero-shot TTS models~\cite{du2024cosyvoice, casanova2024xtts} to isolate the effect of video conditioning. Tab.~\ref{tab:additional_exp}(c) shows that the Lipreading + TTS pipeline is clearly worse than our method in synchronization (LSE-C 3.1/3.7 vs. 7.5; LSE-D 11.4/11.0 vs. 6.8) and content accuracy (WER 31.4/29.4 vs. 20.1), despite competitive audio quality. This is because it generates speech from text alone without using visual dynamics for temporal alignment, further distinguishing our task from conventional TTS.
\paragraph{\textbf{Cross Dataset Validation.}}
Trained on LRS3 and evaluated on the unseen LRS2 in Tab.~\ref{tab:additional_exp}(d), our method achieves the best performance on four of five metrics, including WER (34.9 vs. 39.9) and both synchronization metrics, while ranking second on STOI-Net. These results indicate that our model generalizes more robustly across datasets, preserving both linguistic accuracy and audio-visual synchronization under a domain shift.
\paragraph{\textbf{Robustness of Imperfect Text.}}
Tab.~\ref{tab:imperfect_text} evaluates the framework's resilience against varying levels of textual corruption on the LRS2 dataset. The test set is stratified into three categories based on the lip-reading WER: \textbf{accurate} (WER = 0\%), \textbf{moderately corrupted} ($0\% < \text{WER} \leq 50\%$), and \textbf{severely corrupted} ($\text{WER} > 50\%$).

In terms of textual accuracy (WER), when the auxiliary text is accurate, text-guided models significantly outperform text-free baselines, underscoring the definitive advantage of explicit linguistic conditioning for precise content recovery. More importantly, even under moderate corruption, the WYS framework maintains a substantial WER reduction compared to text-free models. This indicates that our model effectively distills relevant phonetic cues from partially correct text to maintain speech intelligibility, rather than being vulnerable to textual noise that would otherwise lead to a significant increase in WER.

While performance naturally scales with text quality, the LSE-C and DNSMOS scores reveal that our model consistently maintains superior audio-visual synchronization and perceptual quality across all conditions. This confirms that textual conditioning in WYS does not compromise lip-sync fidelity or acoustic naturalness, even when the linguistic guidance is unreliable.

\begin{table}[t!]

\centering
\resizebox{0.9\textwidth}{!}{
\renewcommand{\arraystretch}{1.1}
\begin{tabular}{cc l ccc cc}
  \toprule
  \textbf{WER [\%]} & \multirow{2}{*}{~~\textbf{Ratio}~} & \multirow{2}{*}{~\textbf{Metric}~}
  & \multicolumn{3}{c}{\textbf{Text-free Models}} & \multicolumn{2}{c}{\textbf{Text-guided Models}} \\
  \cmidrule(lr){4-6} \cmidrule(lr){7-8}
  \textbf{(in Lip-reading)} & & & IntelligibleL2S & DiffV2S & V2SFlow-V & LipVoicer & \textbf{Ours (WYS)} \\
  \midrule
  \multirow{3}{*}{\textbf{= 0}} & \multirow{3}{*}{0.61}
  & WER [\%] $\downarrow$ & 25.1919 & 40.6489 & 26.8763 & \textbf{4.2358} & \underline{6.0624} \\
  & & LSE-C $\uparrow$ & 7.1005 & 6.7143 & \underline{7.2917} & 5.9842 & \textbf{7.5834} \\
  & & DNSMOS $\uparrow$ & 2.6625 & 3.0750 & \underline{3.0750} & 3.0529 & \textbf{3.0948} \\
  \midrule
  \multirow{3}{*}{\textbf{(0, 50]}} & \multirow{3}{*}{0.26}
  & WER [\%] $\downarrow$ & 40.5635 & 58.6261 & 40.9356 & \textbf{26.4752} & \underline{28.7612} \\
  & & LSE-C $\uparrow$ & 7.0715 & 6.3118 & \underline{7.2191} & 5.8798 & \textbf{7.5483} \\
  & & DNSMOS $\uparrow$ & 2.6683 & 2.9444 & \textbf{3.1376} & 3.0912 & \underline{3.1134} \\
  \midrule
  \multirow{3}{*}{\textbf{$>$ 50}} & \multirow{3}{*}{0.13}
  & WER [\%] $\downarrow$ & \textbf{75.2857} & 89.4285 & 78.0021 & \underline{78.5714} & 78.7142 \\
  & & LSE-C $\uparrow$ & 6.4737 & 5.6688 & \underline{6.5750} & 5.0511 & \textbf{6.9790} \\
  & & DNSMOS $\uparrow$ & 2.5776 & 2.8641 & \textbf{3.0398} & 2.9945 & \underline{3.0273} \\
  \bottomrule
\end{tabular}
}
\vspace{0.4cm}\caption{Robustness evaluation under imperfect text conditions (LRS2). Additional metrics (LSE-D, STOI-Net) are provided in the supplementary material.} \label{tab:imperfect_text}
\end{table}

\vspace{-0.3cm}
\section{Conclusion}
\label{sec:Conclusion}
We proposed \textit{Watch Your Speech} (WYS), a text-aware video-to-speech framework that addresses the inherent one-to-many mapping problem by incorporating explicit textual conditioning from the training phase onward, integrating linguistic constraints through an attention-based embedding fusion module coupled with a conditional flow matching objective for high-fidelity speech generation. Extensive experiments on the LRS2 and LRS3 benchmarks demonstrate that WYS improves audio-visual synchronization (LSE-C/D) %
while maintaining highly competitive textual accuracy (WER), with subjective evaluations further indicating improved perceived naturalness. %
A current limitation is the reliance on lip-reading accuracy at inference. Future work will explore training strategies robust to noisy text and leverage advances in lip-reading models to further improve synthesis quality.

\vspace{-0.4cm}
\section*{Acknowledgements}

This research was supported by G-LAMP Program of the National Research Foundation of Korea (NRF) grant funded by the Ministry of Education (No. RS-2025-25441317), by the  IITP(Institute  of  Information  \&  Communications  Technology Planning \& Evaluation)-ITRC(Information Technology Research Center) grant funded by the Korea government(Ministry of Science and ICT)(IITP-2026-RS-2020-II201602) and by the National Research Foundation of Korea(NRF) grant funded by the Korea government(MSIT) (RS-2025-24533373), by the Cyber Investigation Support Technology Development Program (No. RS-2025-02304983) of the Korea Institute of Police Technology (KIPoT), funded by the Korean National Police Agency.

\bibliography{egbib}

\clearpage
\appendix
\renewcommand{\thesection}{\Alph{section}}
\renewcommand{\thesubsection}{\thesection.\arabic{subsection}}
\renewcommand{\thesubsubsection}{\thesubsection.\arabic{subsubsection}}
\setcounter{table}{0}
\setcounter{figure}{0}
\setcounter{algorithm}{0}
\renewcommand{\thetable}{S\arabic{table}}
\renewcommand{\thefigure}{S\arabic{figure}}
\renewcommand{\thealgorithm}{S\arabic{algorithm}}
\renewcommand{\theHtable}{S\arabic{table}}
\renewcommand{\theHfigure}{S\arabic{figure}}
\renewcommand{\theHalgorithm}{S\arabic{algorithm}}
\begin{center}
    {\LARGE\bfseries Supplementary Material\par}
    \vspace{0.2cm}
    {\Large\bfseries Watch Your Speech: Text-aware Video-to-Speech Synthesis with Textual Conditioning\par}
\end{center}
\vspace{0.5cm}

\section{Algorithm}
Algorithm~\ref{train_algorithm} and Algorithm~\ref{inference_algorithm} present the training and inference phases of the proposed model in detail. During training phase, given a full face video~$\mathbf{X}$, a preprocessing step is first performed. Specifically, a lip-region cropped video, denoted as $\bar{\mathbf{X}}$, is extracted from $\mathbf{X}$.
In addition, a single frame $\mathbf{x}_{t^*}$ is sampled from $\mathbf{X}$. The corresponding ground-truth text $\mathbf{T}_{i\leq t \leq j}$ is also extracted from the sub-sampled video sequence $\mathbf{X}_{p\leq t \leq q}$. Following preprocessing, the model is trained iteratively until the loss converges. At each iteration, an $v$ is sampled from a normal distribution $\mathcal{N}$, and the fused embedding $\mathbf{e}_{\text{fus}}$ is computed through the back-end module, which incorporates both video-text aligned information and visual information. The training objective is optimized using classifier-free guidance loss $\mathcal{L}_{\text{CFG}}$.
Unlike training, the inference phase begins with a full face video $\mathbf{X}$ without speech. The cropped video $\bar{\mathbf{X}}$ and the single frame image $\mathbf{x}_{t^*}$ are obtained in the same manner as in the training phase. However, the corresponding text $\hat{\text{T}}$ is not explicitly given, but is derived from $\mathbf{X}$ through a lip-reading model $\mathcal{G}$. The inference process proceeds in $T$ time steps. The latent variable $\hat{\text{M}}_{T}$ is initialized by sampling from a normal distribution $\mathcal{N}$. Then, the fused embedding $\mathbf{e}_{\text{fus}}$ is computed using the back-end module. Based on this embedding, the velocity field $v_\theta$ is estimated using classifier-free guidance. Through iterative denoising steps, a mel spectrogram is generated. Finally, a vocoder converts the mel spectrogram into a waveform.

\section{Configuration Details}

We investigate the configuration details of the comparative models used in the main paper with those of our proposed methods. As shown in Tab.~\ref{tab:config_table}, most models adopt a fixed learning rate, whereas SVTS~\cite{mira2022svts} employs a scheduling strategy ranging from $1\times10^{-3}$ to $7\times10^{-3}$. The optimizers used across models are either Adam or AdamW. All methods use a consistent sampling rate of 16kHz. Regarding GPU usage, all models are trained on a single GPU, except for V2SFlow~\cite{choi2025v2sflow}, which utilizes 8 GPUs, and LipVoicer~\cite{yemini2023lipvoicer}, which utilizes 4 GPUs. For vocoders, VCA-GAN~\cite{kim2021lip} applies the traditional method Griffin-Lim~\cite{griffin1984signal}, while SVTS~\cite{mira2022svts} uses WaveGAN~\cite{yamamoto2020parallel}. IntelligibleL2S~\cite{choi2023intelligible} follows an end-to-end structure that generates mel spectrograms and employs an internal vocoder. DiffV2S~\cite{Choi_2023_ICCV}, V2SFlow~\cite{choi2025v2sflow}, and the proposed method leverage HiFi-GAN~\cite{kong2020hifi}.

\vfill
\begin{algorithm}[t!]
\caption{Train with Watch Your Speech}\label{train_algorithm}
\renewcommand{\baselinestretch}{1.3} %

\begin{algorithmic}[1]
\STATE Given the full face video $\mathbf{X}$ \\[3pt]
\STATE $\bar{\mathbf{X}} \leftarrow$ Lip-region cropped video from $\mathbf{X}$  \\[3pt]
\STATE $\mathbf{x}_{t^*} \leftarrow$ Single-frame image sampled from $\mathbf{X}$  \\[3pt]
\STATE $\mathbf{T}_{i\leq t \leq j} \leftarrow$ Ground-truth (GT) text aligned with the sub-sampled video $\mathbf{X}_{p\leq t \leq q}$ \\[3pt]
\REPEAT
    \STATE $\epsilon \sim \mathcal{N}(0, \mathbf{I})$ \\[3pt]
    \STATE $\mathbf{e}_{\text{fus}} \leftarrow$ Output generated via the backend module \\[3pt]
    \STATE Classifier-free guided training:  \\[3pt]
        $\mathcal{L}_\text{CFG}~=~\mathbb{E}_{t,\mathbf{M}_0,\mathbf{e}_{\text{fus}},\mathbf{x}_0,\mathbf{x}_1} \left[ \left\| v_\theta \left( {\mathbf{M}_t, t, \mathbf{e}_\text{fus}} \right)-(\mathbf{x}_1-\mathbf{x}_0) \right\|^2 \right]$ \nonumber \\
        $ ~~~~~~~~~~~~~~~~~+~\mathbb{E}_{t,\mathbf{M}_0,v,\mathbf{x}_0,\mathbf{x}_1} \left[ \left\|  v_\theta \left( {\mathbf{M}_t, t,\emptyset} \right) -(\mathbf{x}_1-\mathbf{x}_0) \right\|^2 \right] \nonumber$
\UNTIL converged
\end{algorithmic}
\end{algorithm}

\begin{algorithm}[t!]
\caption{Inference with Watch Your Speech}\label{inference_algorithm}
\renewcommand{\baselinestretch}{1.3} %

\begin{algorithmic}[1]
\STATE Given a full face video $\mathbf{X}$ without speech \\[3pt]
\STATE $\bar{\mathbf{X}} \leftarrow$ Lip-region cropped video from $\mathbf{X}$  \\[3pt]
\STATE $\mathbf{x}_{t^*} \leftarrow$ Single-frame image sampled from $\mathbf{X}$  \\[3pt]
\STATE $\hat{\mathbf{T}} \leftarrow$ Predicted text from $\mathbf{X}$ by lip-reading : \\[3pt] $\hat{\mathbf{T}} = \mathcal{G}(\mathbf{X})$ \\[3pt]
\STATE Initialize $\hat{\mathbf{M}}_T \sim \mathcal{N}(0, \mathbf{I})$ \\[3pt]
\FOR{$t = T$ to $1$}
    \STATE Compute $v_\theta(\hat{\mathbf{M}}_t,t, \mathbf{e}_{\text{fus}})$ \\
        $~~~~~~~= (1+w)\cdot v_\theta(\hat{\mathbf{M}}_t,t,\mathbf{e}_{\text{fus}}) -w\cdot v_\theta(\hat{\mathbf{M}}_t,t, \emptyset)$ \\[3pt]
    \STATE $\mathbf{z}_{T} \sim \mathcal{N}(0,\mathbf{I})$ if $t >1,$ else $\mathbf{z} = 0$ \\[3pt]
    \STATE $ x_{t-\Delta t} = x_t - v_\theta(x_t, t)\Delta t.$ \\[3pt]
\ENDFOR
\STATE Convert mel spectrogram to waveform: \\$\hat{\mathbf{Y}} = \text{Vocoder}(\hat{\mathbf{M}})$ \\[3pt]
\RETURN $\hat{\mathbf{Y}}$
\end{algorithmic}
\end{algorithm}

\begin{table}[t]
\centering

\resizebox{0.80\columnwidth}{!}{
    \renewcommand{\arraystretch}{1.3}
    \begin{tabular}{lcccccc}
      \toprule
      \textbf{Method} & Learning Rate  & Optimizer   & Sampling Rate & Vocoder & GPUs &Params \\
      \cmidrule(lr){1-7}
      VCA-GAN~\cite{kim2021lip} & $1\times10^{-4}$ & Adam  & 16kHz & Griffin-Lim & Single & 50.06M \\
      SVTS~\cite{mira2022svts} & $1\times10^{-3}$ & Adam  & 16kHz & WaveGAN & Single & 87.63M\\
      DiffV2S~\cite{Choi_2023_ICCV} &$1\times10^{-4}$ & AdamW  & 16kHz & HiFi-GAN & Single & 37.54M\\
      IntelligibleL2S~\cite{choi2023intelligible} &$1\times10^{-3}$ & Adam  & 16kHz & End - to - End & Single & 143.79M\\
      LipVoicer~\cite{yemini2023lipvoicer} & $2\times10^{-4}$ & Adam & 16kHz & HiFi-GAN & 4 & 57.24M \\
      V2SFlow~\cite{choi2025v2sflow} &$1\times10^{-3}$ & AdamW  & 16kHz & HiFi-GAN & 8 & 264.97M\\
      WYS (Proposed) &$2\times10^{-4}$ & Adam  & 16kHz & HiFi-GAN & Single & 153.15M\\
      \bottomrule
    \end{tabular}
}\vspace{0.5cm}\caption{Implementation details of the proposed WYS and compared baseline methods.}

\label{tab:config_table}
\end{table}

\begin{table}[t!]

\centering
\resizebox{0.6\textwidth}{!}{
\renewcommand{\arraystretch}{1.0}
\begin{tabular}{cccccc}
  \toprule
  \textbf{Models}
  & IntelligibleL2S
  & DiffV2S
  & V2SFlow-V
  & LipVoicer
  & \textbf{Ours (WYS)}\\
  \midrule
  \textbf{spkSIM} $\uparrow$ & \textbf{0.7089} & 0.5822 & 0.5732 & 0.5713 & \underline{0.6823} \\

  \bottomrule
\end{tabular}
}\vspace{0.5cm}\caption{Speaker similarity evaluation on LRS2.}
\label{tab:speakerID}
\end{table}

\section{Data preprocessing}\label{suppl:Preprocessing}
The data preprocessing involves preparing two types of visual inputs from the original face video $\mathbf{X}$: the Lip-Region Video $\bar{\mathbf{X}}$ and the Speaker Identity Frame $\mathbf{x}_{t^*}$.

\medskip
\noindent
\textbf{Lip-Region Video.} To isolate the lip movements, 68 facial landmarks are first extracted from each frame using the FaceAlignment\footnote{https://github.com/1adrianb/face-alignment} tool. Based on these landmarks, a $96 \times 96$ pixel region centered on the lips is cropped and converted to grayscale. This results in the lip-region video tensor $\bar{\mathbf{X}} \in \mathbb{R}^{M \times H_{\bar{\mathbf{x}}}  \times W_{\bar{\mathbf{x}}} \times C_{\bar{\mathbf{x}}}}$, where $M$ is the number of frames, $H_{\bar{\mathbf{x}}}=W_{\bar{\mathbf{x}}}=96$, and $C_{\bar{\mathbf{x}}}=1$.

\medskip
\noindent
\textbf{Speaker Identity Frame.} For speaker identity, a single frame is randomly sampled from the video. This frame, $\mathbf{x}_{t^*} \in \mathbb{R}^{1 \times H_{{\mathbf{x}}}  \times W_{{\mathbf{x}}} \times C_{{\mathbf{x}}}}$, is resized and maintained in its RGB format, where $H_{\mathbf{x}}=W_{\mathbf{x}}=224$ and $C_{{\mathbf{x}}}=3$.

\begin{figure*}[t!]
\centering
\includegraphics[width=0.65\textwidth]{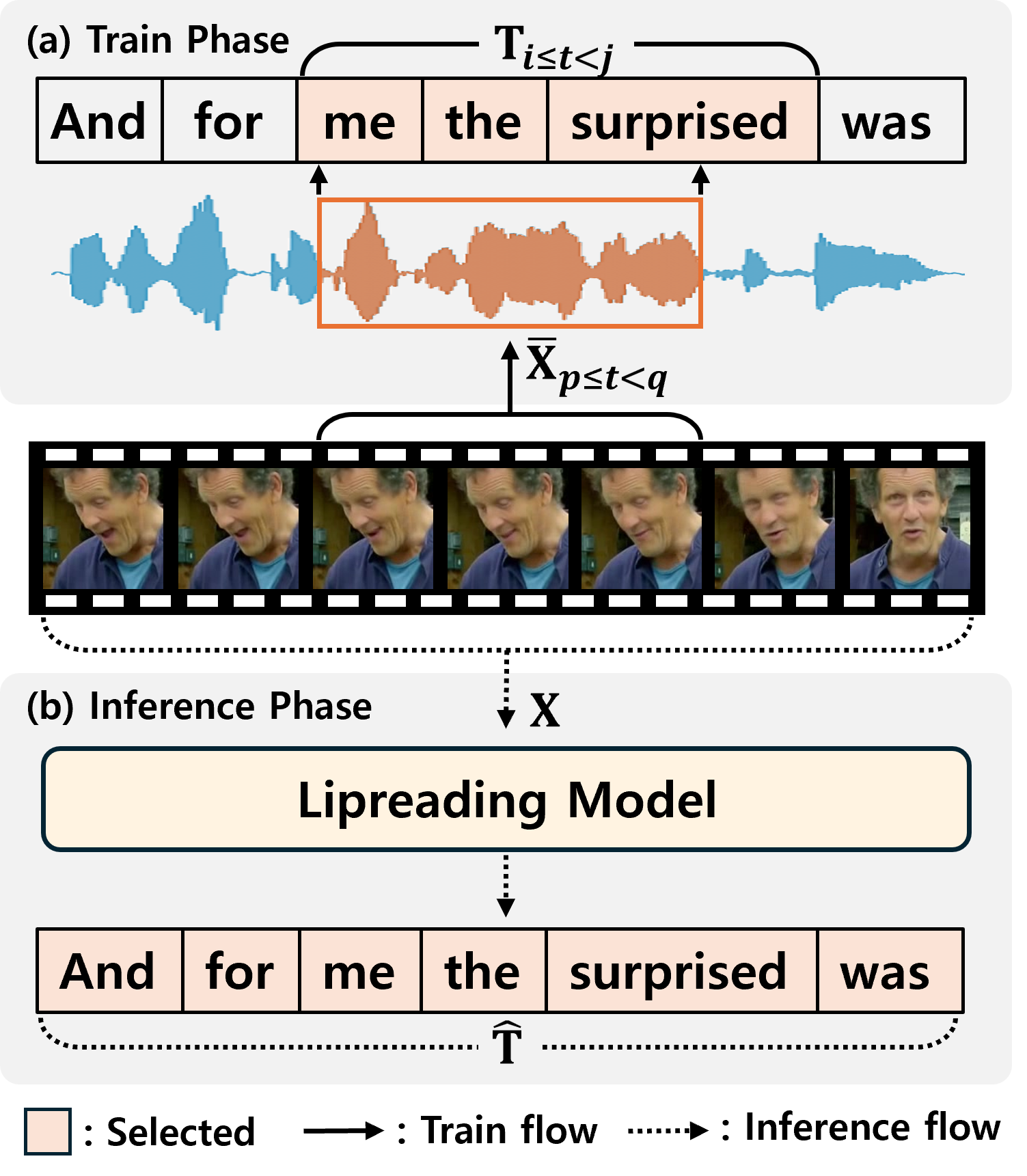}\vspace{0.3cm}
\caption{Illustration of video–text synchronization using word-level timestamps. During training, text segments aligned with the video frames are selected with a small temporal margin. During inference, the full transcript is used as input.
}
\label{fig:text_extract}
\end{figure*}

\medskip
\noindent
\textbf{Temporal Synchronization.} To synchronize video frames with text, we utilize the word-level timestamps provided in the LRS datasets. Fig.~\ref{fig:text_extract} illustrates the process of constructing the text corresponding to given video segment. During training, the words spoken within the temporal range of video frames are selected based on timestamps. If a frame timestamp falls within the duration of a word, the corresponding word is included with a small margin to ensure that the word is fully captured. This process allows the video segment and the text segment to remain temporally aligned. During inference, the entire sentence is used as input without applying this temporal selection process.

\section{Compare Speaker Similarity}
The speaker identity is evaluated using {Speaker Similarity (spkSIM)}. As shown in {Tab.~\ref{tab:speakerID}}, WYS achieves a similarity score of {0.6823}, outperforming all other baselines except IntelligibleL2S. Although our method does not achieve the highest score, it demonstrates competitive performance compared to prior approaches. This indicates that the proposed model maintains a reasonable level of speaker identity while focusing primarily on generating natural speech from visual inputs.

\section{Lip-Reading Backbone Selection}
Tab.~\ref{tab:lipreading_benchmark} shows the WER comparison on the LRS2 and LRS3 datasets. Auto-AVSR~\cite{ma2023auto} achieves lower WER than AV-Hubert~\cite{shi2022avhubert} on both datasets, demonstrating strong lip-reading performance. Therefore, we select Auto-AVSR as the backbone for our experiments.

\begin{table}[t!]

\centering
\resizebox{0.5\textwidth}{!}{
\renewcommand{\arraystretch}{1.0}
\begin{tabular}{ccc}
  \toprule
  \textbf{Models} & Auto-AVSR & AV-Hubert \\
  \midrule
  \textbf{LRS2 / LRS3} & \textbf{14.60 / 19.10} & 23.82 / 25.51 \\
  \bottomrule
\end{tabular}
}\vspace{0.5cm}\caption{WER results for lip-reading backbone benchmarks.}
\label{tab:lipreading_benchmark}
\end{table}

\begin{table}[t!]

\centering
\resizebox{0.8\textwidth}{!}{
\renewcommand{\arraystretch}{1.1}
\begin{tabular}{cc l ccc cc}
  \toprule
  \textbf{WER [\%]} & \multirow{2}{*}{~~\textbf{Ratio}~} & \multirow{2}{*}{~\textbf{Metric}~}
  & \multicolumn{3}{c}{\textbf{Text-free Models}} & \multicolumn{2}{c}{\textbf{Text-guided Models}} \\
  \cmidrule(lr){4-6} \cmidrule(lr){7-8}
  \textbf{(in Lip-reading)} & & & IntelligibleL2S & DiffV2S & V2SFlow-V & LipVoicer & \textbf{Ours (WYS)} \\
  \midrule
  \multirow{5}{*}{\textbf{= 0}} & \multirow{5}{*}{0.61}
  & WER [\%] $\downarrow$ & 25.1919 & 40.6489 & 26.8763 & \textbf{4.2358} & \underline{6.0624} \\
  & & LSE-C $\uparrow$ & \underline{7.1005} & 6.7143 & \underline{7.2917} & 5.9842 & \textbf{7.5834} \\
  & & LSE-D $\downarrow$ & 7.0827 & 7.5199 & 7.1889 & 8.2879 & \textbf{6.7649} \\
  & & STOI-Net $\uparrow$ & 0.8594 & 0.8931 & \textbf{0.9210} & 0.9048 & \underline{0.9137} \\
  & & DNSMOS $\uparrow$ & 2.6625 & 3.0750 & \underline{3.0750} & 3.0529 & \textbf{3.0948} \\
  \midrule
  \multirow{5}{*}{\textbf{(0, 50]}} & \multirow{5}{*}{0.26}
  & WER [\%] $\downarrow$ & 40.5635 & 58.6261 & 40.9356 & \textbf{26.4752} & \underline{28.7612} \\
  & & LSE-C $\uparrow$ & 7.0715 & 6.3118 & \underline{7.2191} & 5.8798 & \textbf{7.5483} \\
  & & LSE-D $\downarrow$ & \underline{7.1249} & 7.7609 & 7.3305 & 8.4246 & \textbf{6.8357} \\
  & & STOI-Net $\uparrow$ & 0.8550 & 0.8912 & \textbf{0.9227} & 0.9073 & \underline{0.9136} \\
  & & DNSMOS $\uparrow$ & 2.6683 & 2.9444 & \textbf{3.1376} & 3.0912 & \underline{3.1134} \\
  \midrule
  \multirow{5}{*}{\textbf{$>$ 50}} & \multirow{5}{*}{0.13}
  & WER [\%] $\downarrow$ & \textbf{75.2857} & 89.4285 & 78.0021 & \underline{78.5714} & 78.7142 \\
  & & LSE-C $\uparrow$ & 6.4737 & 5.6688 & \underline{6.5750} & 5.0511 & \textbf{6.9790} \\
  & & LSE-D $\downarrow$ & \underline{7.2273} & 8.2211 & 7.5075 & 8.7748 & \textbf{6.9457} \\
  & & STOI-Net $\uparrow$ & 0.8397 & 0.8903 & \textbf{0.9233} & 0.8998 & \underline{0.9126} \\
  & & DNSMOS $\uparrow$ & 2.5776 & 2.8641 & \textbf{3.0398} & 2.9945 & \underline{3.0273} \\
  \bottomrule
\end{tabular}
}\vspace{0.5cm}\caption{Robustness evaluation under imperfect text conditions (LRS2).}
\label{tab:full_imperfect_text}
\end{table}

\section{Robustness to Imperfect Text}
Tab.~\ref{tab:full_imperfect_text} provides a more detailed analysis under imperfect textual conditions on the LRS2 dataset, reporting LSE-D and STOI-Net in addition to the metrics discussed in the main paper. The additional metrics exhibit trends consistent with the main results. In particular, WYS maintains low LSE-D values across all text corruption levels, indicating stable audio–visual synchronization. For STOI-Net, text-free models achieve slightly higher scores in some conditions, while WYS remains competitive overall. We also observe that as text corruption increases, WER increases significantly across all models, whereas STOI-Net and DNSMOS change more gradually. Overall, WYS maintains a balanced performance between linguistic accuracy and perceptual speech quality under varying levels of textual reliability.

\section{Qualitative Comparison of Mel Spectrograms}

\subsection{Comparison with Other Models}
\begin{figure*}[t!]
\centering
\includegraphics[width=0.8\textwidth]{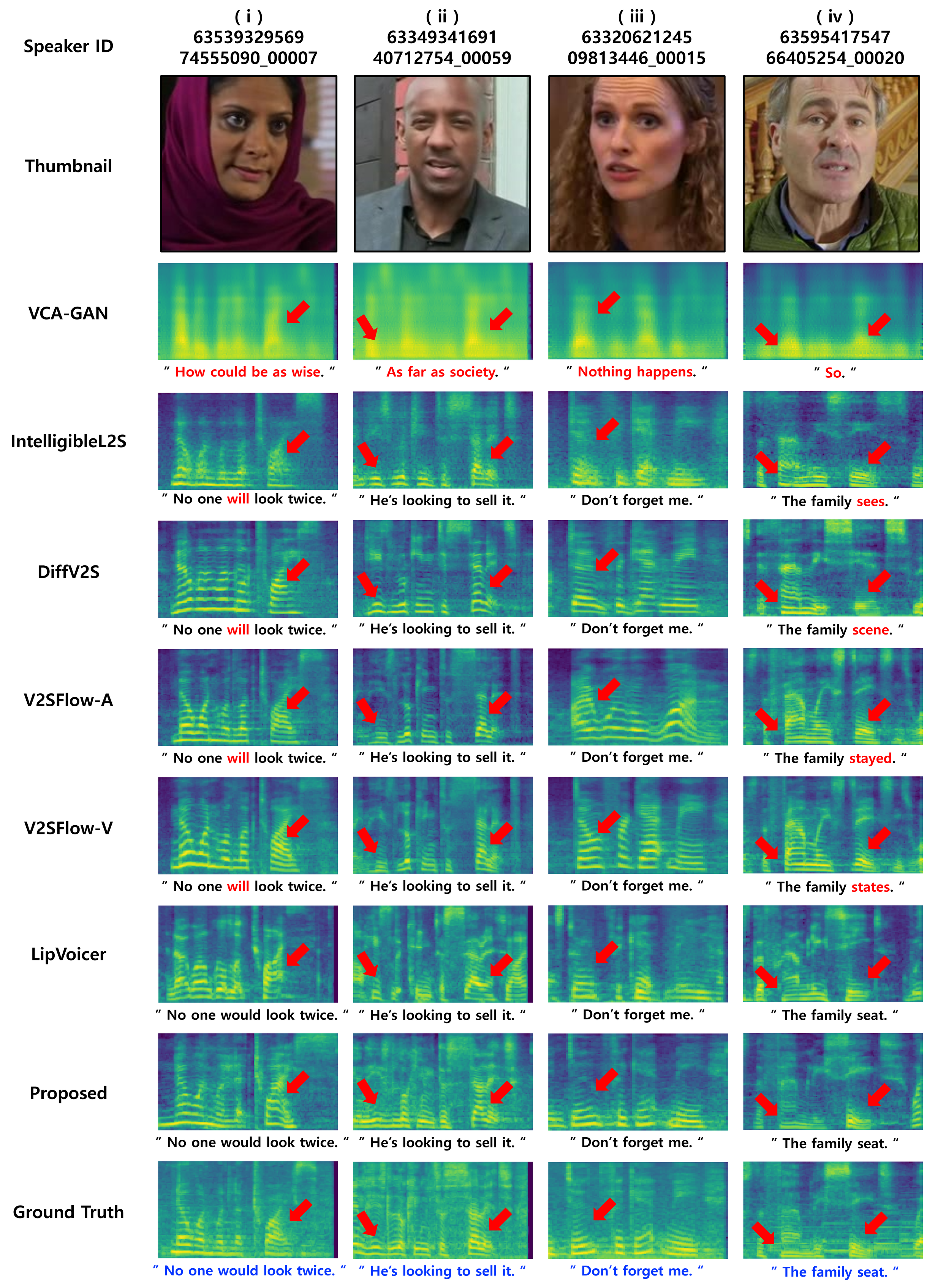}\vspace{0.3cm}
\caption{Qualitative comparison of mel spectrograms for LRS2 dataset. The scripts below each mel spectrogram represent the \textbf{ASR-predicted text} in \textbf{black}, while the \textcolor{blue}{ground-truth (GT) text} is shown beneath the GT mel spectrogram in \textcolor{blue}{blue}. \textcolor{red}{Incorrectly predicted words} are highlighted in \textcolor{red}{red}.
}
\label{fig:mel_lrs2}
\end{figure*}

\begin{figure*}[t!]
\centering
\includegraphics[width=0.8\textwidth]{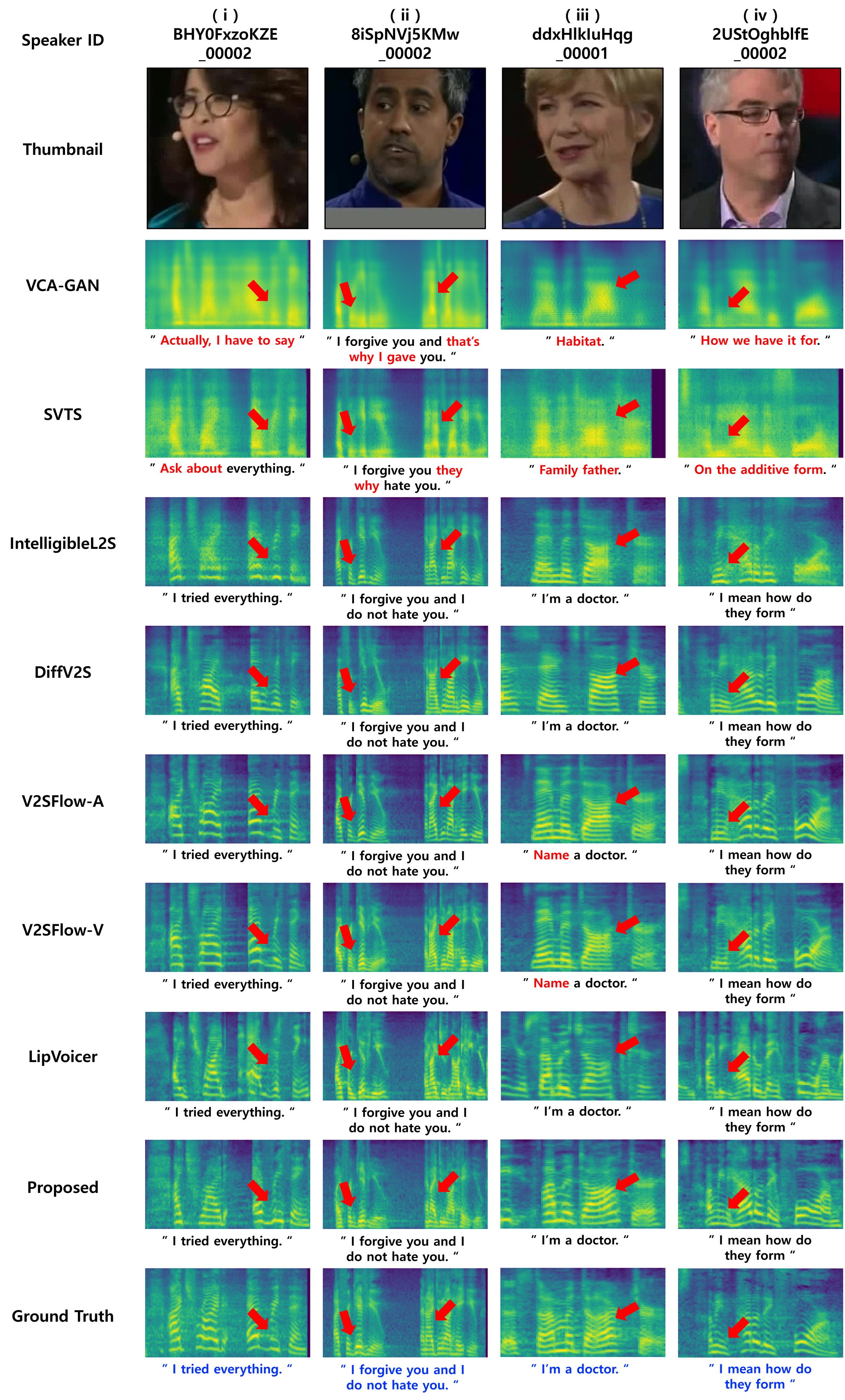}\vspace{0.3cm}
\caption{Qualitative comparison of mel spectrograms for LRS3 dataset. The scripts below each mel spectrogram represent the \textbf{ASR-predicted text} in \textbf{black}, while the \textcolor{blue}{ground-truth (GT) text} is shown beneath the GT mel spectrogram in \textcolor{blue}{blue}. \textcolor{red}{Incorrectly predicted words} are highlighted in \textcolor{red}{red}.
}
\label{fig:mel_lrs3}
\end{figure*}

We present a qualitative comparison of the mel spectrograms generated by the comparative models and our proposed method. As illustrated in Fig.~\ref{fig:mel_lrs2} and Fig.~\ref{fig:mel_lrs3}, our method generates mel spectrograms that closely match the ground-truth, capturing detailed acoustic features and harmonic patterns across diverse samples from the LRS2 and LRS3 dataset. In contrast, VCA-GAN~\cite{kim2021lip}, SVTS~\cite{mira2022svts}, and DiffV2S~\cite{Choi_2023_ICCV} generate blurred mel spectrograms compared to ground-truth (GT), which often results in inaccurate speech generation. IntelligibleL2S~\cite{choi2023intelligible} and V2SFlow~\cite{choi2025v2sflow} show closer approximations, but still exhibit differences in fine details compared to ground truth (GT). On the other hand, our proposed model accurately preserves subtle spectral features and visually aligns best with the original mel spectrograms. This fidelity in mel spectrograms contributes to the generation of speech that is both intelligible and natural. We further validate this observation through Automatic-Speech-Recognition(ASR) evaluation, which confirms that our proposed model consistently achieves the most stable performance across samples.

\subsection{Failure Case}
\begin{figure*}[t!]
\centering
\includegraphics[width=0.8\textwidth]{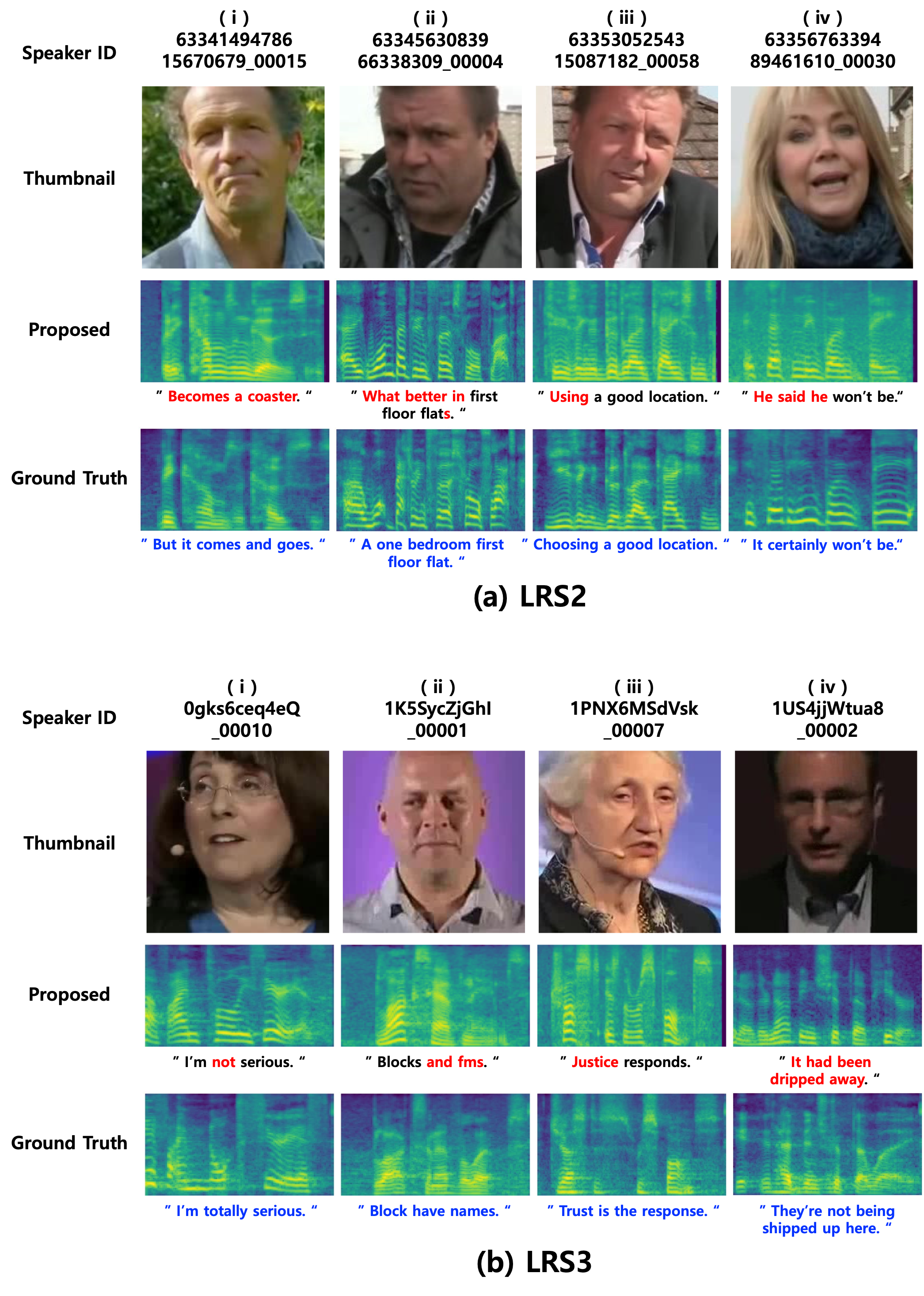}\vspace{0.3cm}
\caption{Failure cases in mel spectrogram generation with incorrect text predictions on (a) LRS2 and (b) LRS3. The scripts below each mel spectrogram represent the \textbf{ASR-predicted text} in \textbf{black}, while the \textcolor{blue}{ground-truth (GT) text} is shown beneath the GT mel spectrogram in \textcolor{blue}{blue}. \textcolor{red}{Incorrectly predicted words} are highlighted in \textcolor{red}{red}.
}
\label{fig:failure_case}
\end{figure*}

Fig.~\ref{fig:failure_case} presents failure cases where the original audio is not accurately reconstructed on LRS2 and LRS3. Although the proposed model achieves superior performance in terms of Word Error Rate (WER) compared to comparative models, it still fails to fully capture the correct textual content in certain instances. This limitation arises from the use of pseudo-text generated by the lip-reading model during inference, which imposes an upper bound on performance. Since the quality of the guided embedding depends on the accuracy of the lipreading output, errors in the pseudo-text can lead to suboptimal reconstructions.

\subsection{Effect of Attention Mechanism}
\begin{figure*}[t!]
\centering

\begin{minipage}[t]{1.0\textwidth}
    \centering
    \includegraphics[width=0.8\textwidth]{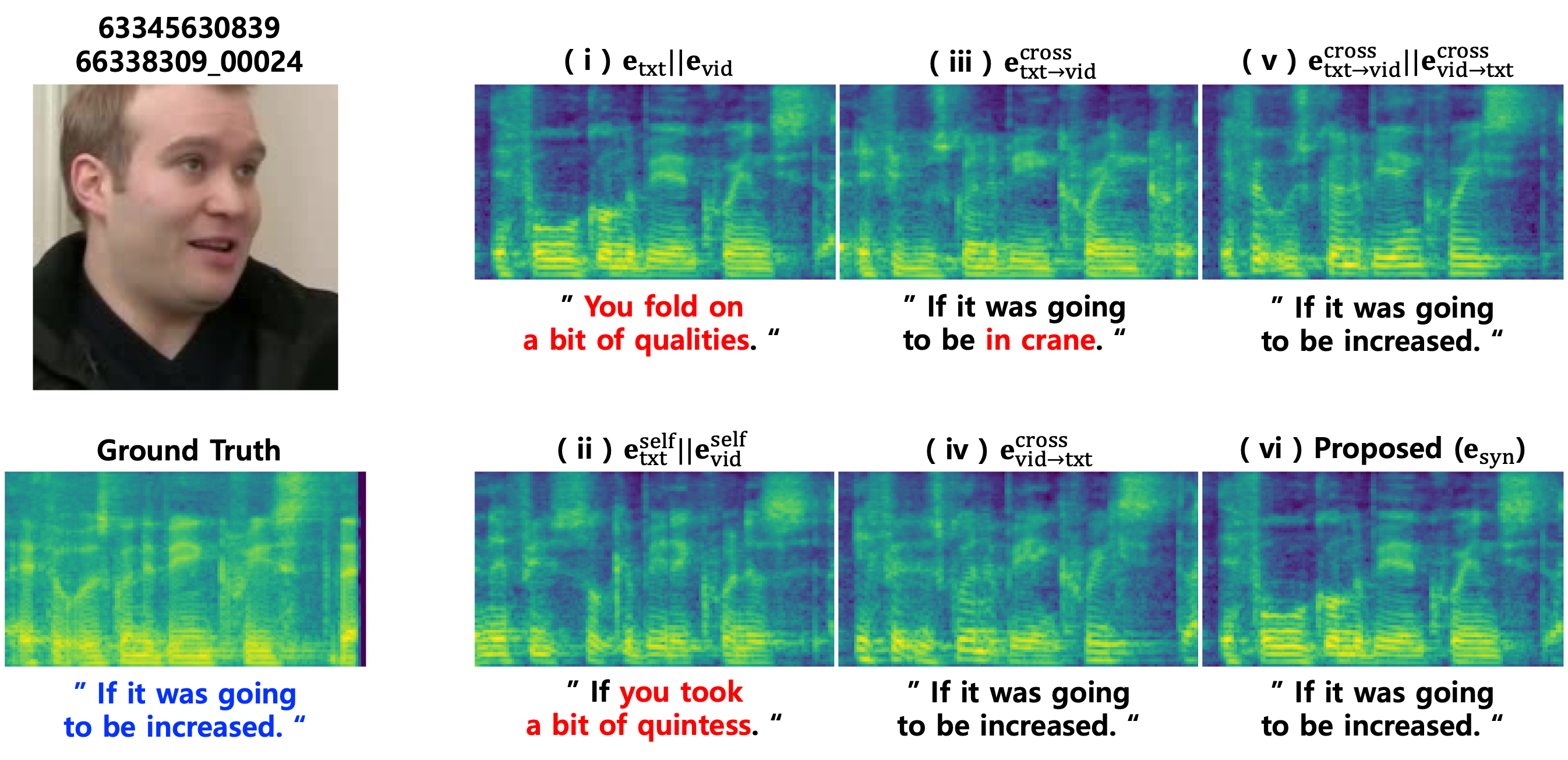}\vspace{0.3cm}
    \captionof{figure}{Qualitative comparison of mel spectrograms based on different embedding combinations in the attention mechanism for the LRS2 Dataset. The scripts below each mel spectrogram represent the \textbf{ASR-predicted text} in \textbf{black}, while the \textcolor{blue}{ground-truth (GT) text} is shown beneath the GT mel spectrogram in \textcolor{blue}{blue}. \textcolor{red}{Incorrectly predicted words} are highlighted in \textcolor{red}{red}.
}
    \label{fig:mel_attention}
\end{minipage}%
\hfill
\vspace{2em}
\begin{minipage}[t]{0.475\textwidth}
    \centering
    \includegraphics[width=0.65\textwidth]{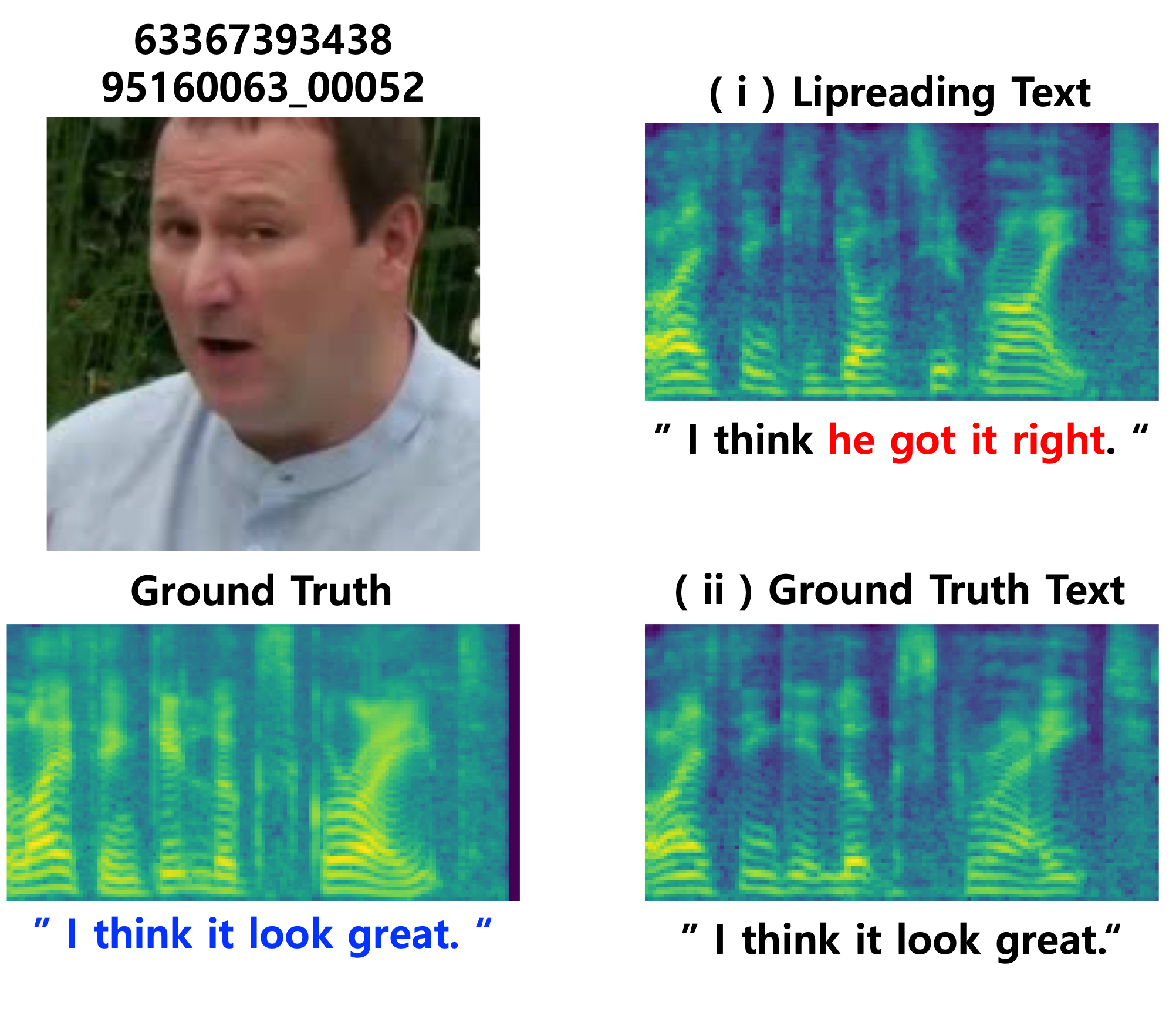}\vspace{0.3cm}
    \captionof{figure}{Mel spectrogram comparison on LRS2 using different guiding texts during inference: (i) Lipreading Text, (ii) Ground-Truth Text. The scripts below each mel spectrogram represent the \textbf{ASR-predicted text} in \textbf{black}, while the \textcolor{blue}{ground-truth (GT) text} is shown beneath the GT mel spectrogram in \textcolor{blue}{blue}. \textcolor{red}{Incorrectly predicted words} are highlighted in \textcolor{red}{red}.
 }
    \label{fig:mel_lipreading}
\end{minipage}%
\hfill
\begin{minipage}[t]{0.475\textwidth}
    \centering
    \includegraphics[width=1.0\textwidth]{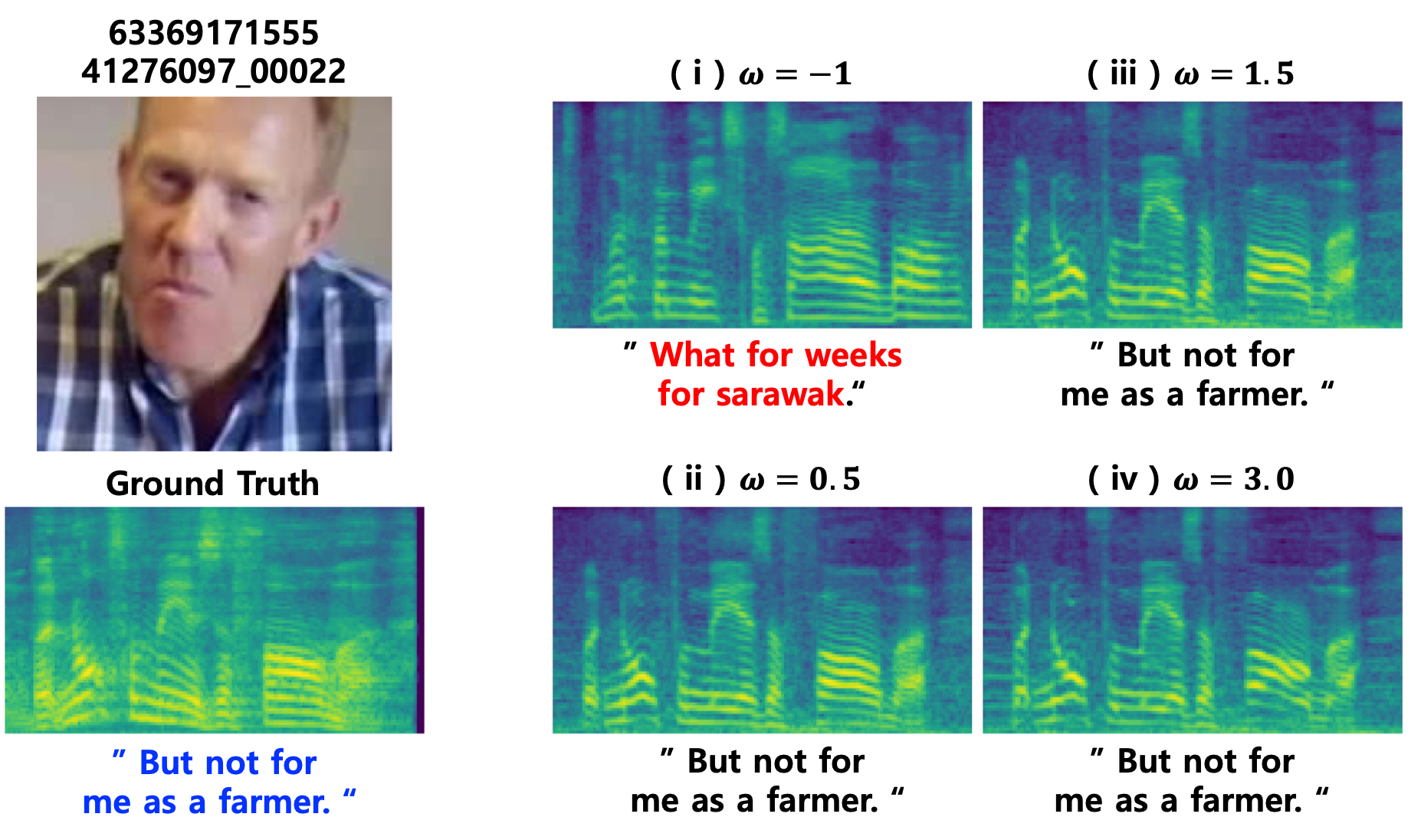}\vspace{0.3cm}
    \captionof{figure}{Mel spectrogram comparison on LRS2 with different values of the guidance scaling factor $\omega$. The scripts below each mel spectrogram represent the \textbf{ASR-predicted text} in \textbf{black}, while the \textcolor{blue}{ground-truth (GT) text} is shown beneath the GT mel spectrogram in \textcolor{blue}{blue}. \textcolor{red}{Incorrectly predicted words} are highlighted in \textcolor{red}{red}.
}
    \label{fig:mel_scale}
\end{minipage}

\end{figure*}

Fig.~\ref{fig:mel_attention} compares the mel spectrograms generated under different configurations of the attention mechanism. $\mathbf{e}_{\text{m}}$ and $\mathbf{e}_{\text{m}}^\text{self}$ denote the encoder output and the self-attention output, respectively, where $m \in \{ \text{txt}, \text{vid} \} $ indicates the modality (i.e., text or video). $\mathbf{e}_{m \to w}^\text{cross}$ is the cross-attention output when modality $m$ serves as the query. $\mathbf{e}_\text{syn}$ is the proposed embedding. As shown in Fig.~\ref{fig:mel_attention} (i-iii), the resulting mel spectrograms fail to properly reflect the underlying text information, indicating the limitations of these configurations. In contrast, Fig.~\ref{fig:mel_attention} (iv-vi), which incorporate attention mechanisms, demonstrate more accurate guidance from attention mechanism. Notably, our proposed model in Fig.~\ref{fig:mel_attention} (vi) yields a mel spectrogram that most closely resembles the ground truth, particularly in the low-frequency regions. These results substantiate the effectiveness of the proposed approach.

\subsection{Effect of Lip-Reading Model}
We compare mel spectrograms generated using different text inputs during inference. In Fig.~\ref{fig:mel_lipreading} (i), the proposed method utilizes pseudo-text generated by lipreading model as the input. In contrast, Fig.~\ref{fig:mel_lipreading} (ii) shows the result when the GT text is provided instead. A comparison of the text outputs reveals that using GT text as input yields higher text accuracy. Furthermore, mel spectrogram with GT text input exhibits greater similarity to the GT mel spectrogram, compared to that generated using lipreading based pseudo-text.

\subsection{Effect of Guidance Scale Factor}
Fig.~\ref{fig:mel_scale} illustrates the effect of the guidance scale factor $\omega$ on the resulting mel spectrograms. As shown in Fig.~\ref{fig:mel_scale} (ii-iv), when guidance is applied, the model generates stable and coherent mel spectrograms that closely resemble the GT mel spectrogram. In contrast, Fig.~\ref{fig:mel_scale} (i), which corresponds to the case without guidance, exhibits a distorted and unstable mel spectrogram, highlighting the importance of the guidance.

\subsection{Effect of Sampling Steps}
\begin{figure*}[t!]
\centering
\includegraphics[width=0.8\textwidth]{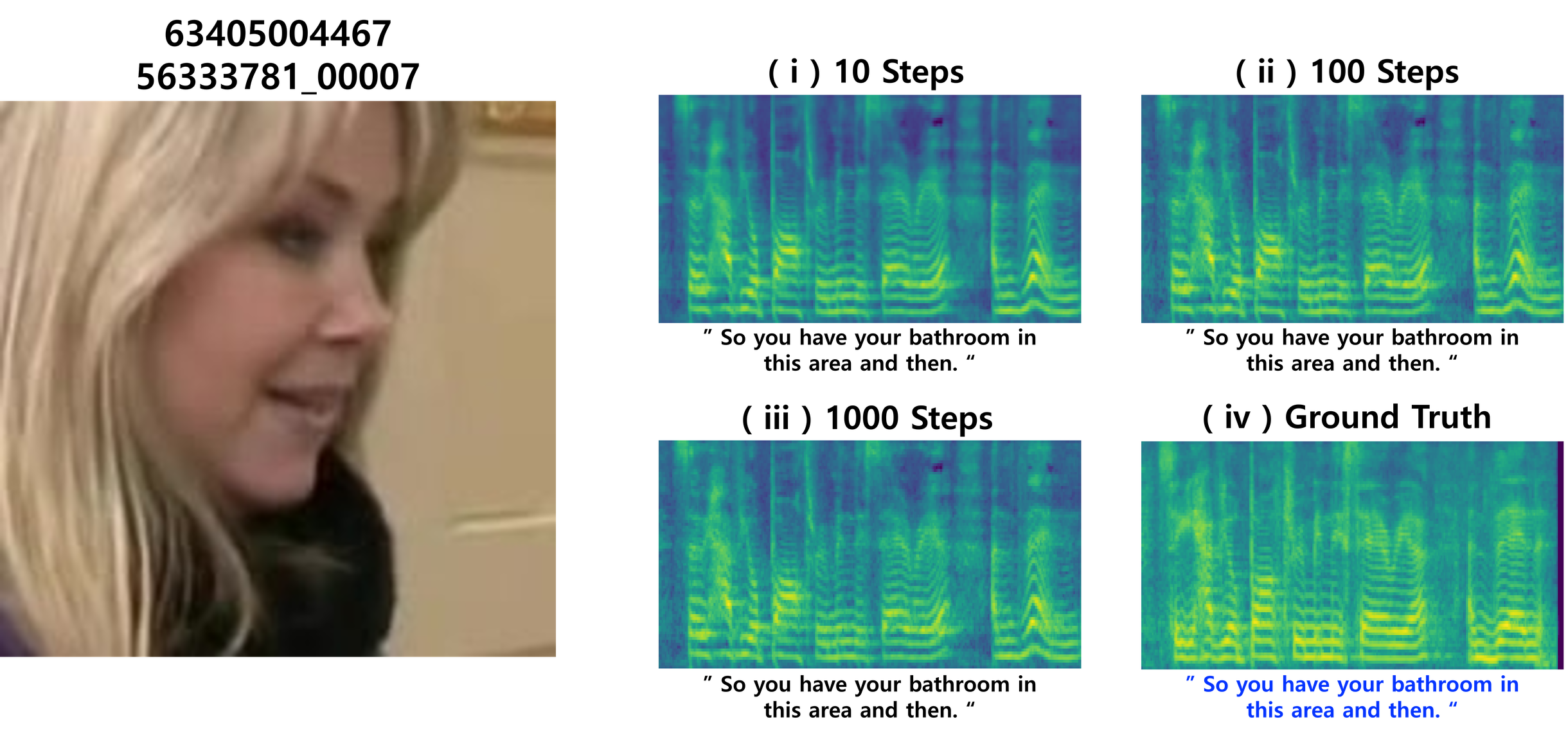}\vspace{0.3cm}
\caption{Mel spectrogram comparison on LRS2 using different
sampling steps: (i) 10 steps, (ii) 100 steps, (iii) 1000 steps, and (iv) Ground-Truth. The scripts below each mel spectrogram represent the
ASR-predicted text in black, while the ground-truth (GT) text
is shown beneath the GT mel spectrogram in blue.}
\label{fig:mel_steps}
\end{figure*}

Fig.~\ref{fig:mel_steps} shows the mel spectrograms generated with different sampling steps. As observed in Fig.~\ref{fig:mel_steps} (i-iii), the overall structure of the mel spectrograms remains largely unchanged across varying sampling steps. Moreover, even with as few as 10 steps, the model produces perceptually meaning mel spectrograms. While the baseline setup of the original paper employs 100 steps, our results suggest that using only 10 steps can be an effective and efficient alternative in practical applications.

\section{Crowdsource Evaluation}

\begin{figure*}[t!]
\centering
\includegraphics[width=1.0\textwidth]{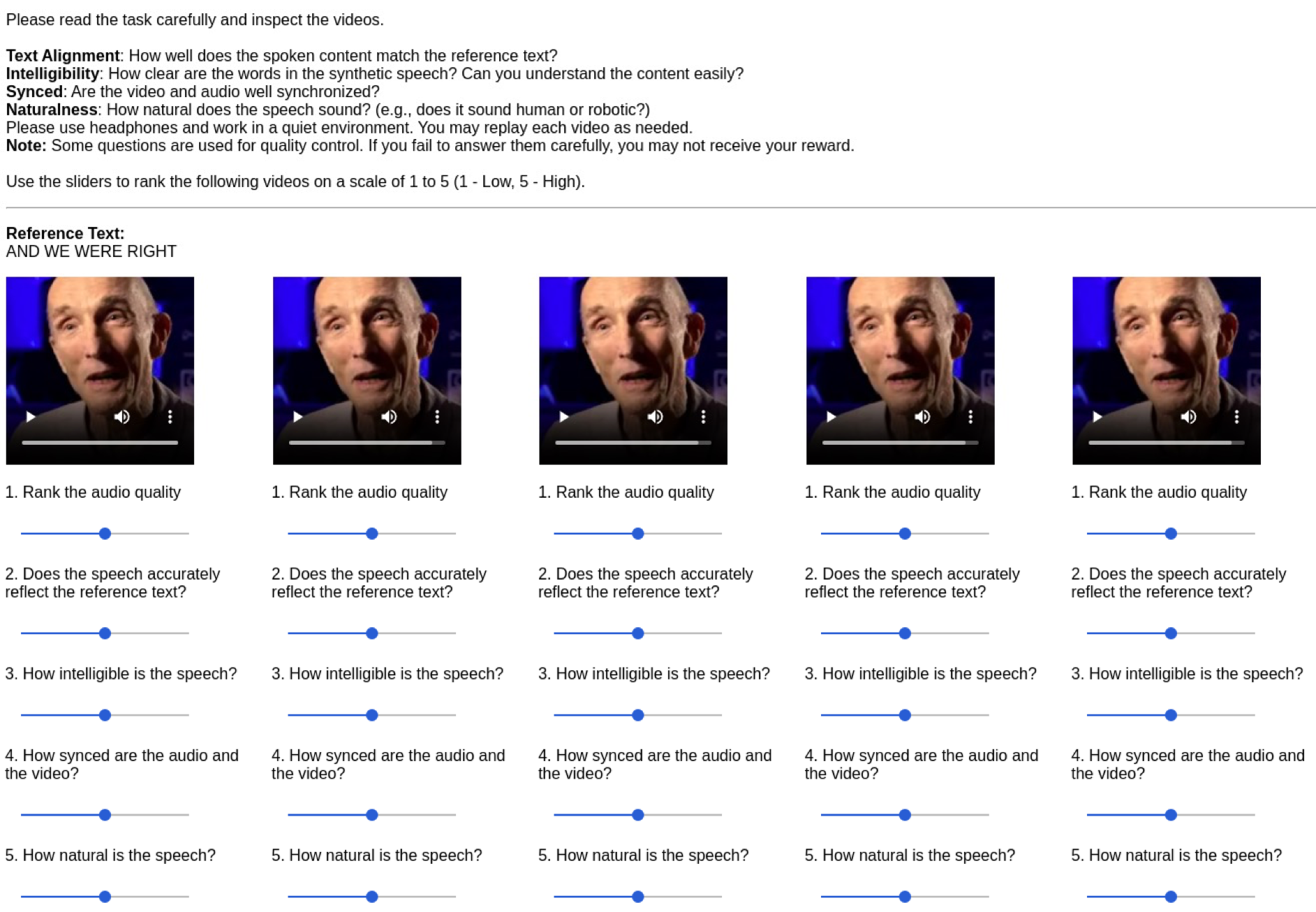}\vspace{0.3cm}
\caption{MOS evaluations. Participants evaluated samples from IntelligibleL2S \cite{choi2023intelligible}, DiffV2S \cite{Choi_2023_ICCV}, LipVoicer \cite{yemini2023lipvoicer}, V2SFlow-V \cite{choi2025v2sflow}, WYS (Ours), and the ground-truth based on five criteria, with randomized sample.}
\label{fig:survey}
\end{figure*}

Fig.~\ref{fig:survey} presents the instructions displayed on the evaluation webpage used for crowdsourced MOS assessments via Amazon Mechanical Turk (MTurk). The evaluation criteria include overall audio fidelity (quality), correspondence to the reference text (alignment), clarity of the spoken content (intelligibility), audio-visual consistency (synchronization), and perceived human-likeness of the speech (naturalness), as illustrated in Fig.~\ref{fig:survey}.
On MTurk, each task unit is referred to as a Human Intelligence Task (HIT). For the MOS evaluation, a total of 20 HITs were created, with 10 for the LRS2 dataset and 10 for the LRS3 dataset. Each HIT required participants to evaluate five video samples corresponding to different models and was assigned to 10 unique participants, with a reward of 0.5 USD per HIT.

\end{document}